\documentclass[lettersize,journal]{IEEEtran}
\usepackage{amsmath,amsfonts}
\usepackage{algorithmic}
\usepackage{algorithm}
\usepackage{array}
\usepackage[caption=false,font=normalsize,labelfont=sf,textfont=sf]{subfig}
\usepackage{textcomp}
\usepackage{stfloats}
\usepackage{url}
\usepackage{verbatim}
\usepackage{graphicx}
\usepackage{cite}
\usepackage{multirow} 
\usepackage{booktabs} 
\usepackage{graphicx} 
\usepackage{xcolor}

\begin{document}

\title{Immunizing 3D Gaussian Generative Models Against Unauthorized Fine-Tuning via Attribute-Space Traps}

\author{Jianwei Zhang,
          Sihan Cao,
          Chaoning Zhang,~\IEEEmembership{Senior Member,~IEEE,}
          Ziming Hong,
          Jiaxin Huang,~\IEEEmembership{Graduate Student Member,~IEEE,}
          Pengcheng Zheng,
          Caiyan Qin,
          Wei Dong,
          Yang Yang,~\IEEEmembership{Senior Member,~IEEE,}
          and Tongliang Liu,~\IEEEmembership{Senior Member,~IEEE}
\thanks{This work was partially supported by the National Natural Science Foundation of China under Grant 62572104 and Grant 62220106008.}%
  \thanks{Jianwei Zhang, Sihan Cao, Chaoning Zhang, Pengcheng Zheng and Yang Yang are with the School of Computer Science and Engineering, University of
  Electronic Science and Technology of China, Chengdu 611731, China (e-mail: zjw5428c@gmail.com; 2023080903002@std.uestc.edu.cn;
  chaoningzhang1990@gmail.com; zpc777@std.uestc.edu.cn; yang.yang@uestc.edu.cn).}%
  \thanks{Ziming Hong and Tongliang Liu are with the Sydney AI Centre, The University of Sydney, Sydney, NSW 2006, Australia (e-mail:
  hoongzm@gmail.com; tongliang.liu@sydney.edu.au).}%
  \thanks{Jiaxin Huang is with the Machine Learning Department, Mohamed bin Zayed University of Artificial Intelligence, Abu Dhabi, United
  Arab Emirates (e-mail: jiaxin.huang@mbzuai.ac.ae).}%
  \thanks{Caiyan Qin is with the School of Robotics and Advanced Manufacture, Harbin Institute of Technology, Shenzhen 518055, China (e-mail:
  qincaiyan@hit.edu.cn).}%
  \thanks{Wei Dong is with the School of Information and Control Engineering, Xi'an University of Architecture and Technology, Xi'an 710055,
  China (e-mail: dongwei156@xauat.edu.cn).}%
  }

\markboth{SUBMITTED TO IEEE TRANSACTIONS ON MULTIMEDIA}%
  {Zhang \MakeLowercase{\textit{et al.}}}


\maketitle

\begin{abstract}

Recent large-scale generative models enable high-quality 3D synthesis. However, the public accessibility of pre-trained weights introduces a critical vulnerability. Adversaries can fine-tune these models to steal specialized knowledge acquired during pre-training, leading to intellectual property infringement. Unlike defenses for 2D images and language models, 3D generators require specialized protection due to their explicit Gaussian representations, which expose fundamental structural parameters directly to gradient-based optimization. We propose GaussLock, the first approach designed to defend 3D generative models against fine-tuning attacks. GaussLock is a lightweight parameter-space immunization framework that integrates authorized distillation with attribute-aware trap losses targeting position, scale, rotation, opacity, and color. Specifically, these traps systematically collapse spatial distributions, distort geometric shapes, align rotational axes, and suppress primitive visibility to fundamentally destroy structural integrity. By jointly optimizing these dual objectives, the distillation process preserves fidelity on authorized tasks while the embedded traps actively disrupt unauthorized reconstructions. Experiments on large-scale Gaussian models demonstrate that GaussLock effectively neutralizes unauthorized fine-tuning attacks. It substantially degrades the quality of unauthorized reconstructions, evidenced by significantly higher LPIPS and lower PSNR, while effectively maintaining performance on authorized fine-tuning.
\end{abstract}


\section{Introduction}
\label{sec:intro}
\IEEEPARstart{L}{arge-scale} 3D generative models~\cite{achlioptas2018learning,chaudhuri2020learning,gao2022get3d,geiger2011stereoscan,sayed2022simplerecon,shi2022deep,wang2023rodin,xu2024grm,wen20233d,zheng2026llava} are becoming a key infrastructure for modern multimedia content creation. By enabling the efficient synthesis of high-fidelity 3D assets, they substantially reduce production costs for applications like games~\cite{buyuksalih20173d} and interactive media~\cite{cellary2012interactive}. However, training these high-quality generators requires massive multi-view data and considerable computation. The resulting pre-trained weights embody substantial structured knowledge and engineering investment. As these weights are increasingly released, an urgent security concern emerges. An adversary can perform unauthorized downstream adaptation (e.g., fine-tuning)~\cite{zhang2025purifier} with a few target data to produce a derivative model, thereby stealing proprietary 3D structure and generation capabilities~\cite{chattopadhyay2025one,dolgavin2025turning,zheng2025joint}. We focus on this white-box fine-tuning threat, where an attacker leverages accessible pre-trained weights and a small amount of target data
  to directly and effectively transfer the underlying 3D structural and generative priors of the model.

Several trends in 3D generative modeling~\cite{gao2022get3d,mildenhall2021nerf,schwarz2020graf,zhang2026learning} make this fine-tuning capability theft particularly feasible. The objective of unauthorized adaptation in 3D is to transfer geometric structure and attribute regularities that remain consistent across viewpoints. First, an increasing number of 3D pipelines adopt explicit and editable representations like Gaussian primitives~\cite{qin2024langsplat,yi2024gaussiandreamer,ren2024l4gm,zhu2025large,tang2024lgm}, where geometry and appearance attributes act as physical parameters. For instance, feed-forward Gaussian generators~\cite{wang2025f3d,yang2025prometheus,chen2024pref3r,zheng2025lightweight} like LGM directly output explicit attribute vectors such as position, scale, rotation, opacity, and color, providing a natural interface for optimization. Second, multi-view consistency and differentiable rendering provide strong supervisory signals that enable rapid overfitting and structural transfer. Third, parameter-efficient adaptation lowers the barrier to producing high-quality derivatives. Ultimately, explicit attribute interfaces, strong multi-view supervision, and low-cost adaptation elevate the risk that a small target dataset can yield a usable derivative model. This motivates defenses that directly constrain how 3D structures become consolidated.

Despite recent progress in protecting 2D vision ~\cite{sha2022fine,wang2021non,deng2024sophonnonfinetunablelearningrestrain,hong2025toward} and language models~\cite{rosati2024representationnoisingdefencemechanism,chen2025sddselfdegradeddefensemalicious,huang2024harmful}, effective and lightweight defenses remain limited for 3D generative models based on explicit Gaussian representations. 3D content must satisfy cross-view geometric consistency and physical plausibility. Because structural and attribute information is exposed in a parameterized form, an attacker can consolidate stolen geometric regularities directly in the parameter space. Consequently, defenses acting only on rendered outputs struggle to reliably constrain the key variables that determine 3D structure. A usable derivative model in this context generates 3D content with cross-view consistency rather than just superficially similar single-view renderings. These observations call for tailored defenses that suppress structural capability transfer at the parameter level while preserving original utility with minimal overhead. 

To address this problem, we propose GaussLock, the first approach designed to defend 3D generative models against fine-tuning attacks. We define the source domain $\mathcal{D}_{src}$ as the authorized data distribution used for pre-training and intended tasks, and the target domain $\mathcal{D}_{tgt}$ as the specific unauthorized data an attacker uses to fine-tune the model. By utilizing only a few samples from $\mathcal{D}_{tgt}$, an attacker can effectively fit the model to steal the specialized knowledge and proprietary structures acquired during pre-training. Our core insight is to embed dormant traps within the physical attributes of the Gaussian representation. These traps remain inactive during normal inference on $\mathcal{D}_{src}$ but trigger a structural collapse if unauthorized optimization is detected on $\mathcal{D}_{tgt}$. GaussLock operates directly in the parameter space instead of the 2D rendering space because a parameter-level defense is invariant to camera viewpoints. To ensure the original utility of the generator on $\mathcal{D}_{src}$ is maintained, we incorporate a parameter-space source distillation strategy. Empirically, our experiments demonstrate that such parameter-level physical traps successfully induce structural collapse on unauthorized target domains, while strictly maintaining high-fidelity generation on the original source tasks.

To conclude, the main contributions of this paper are
summarized as follows:
\begin{itemize}
    \item We present the first study on defending 3D generative models against fine-tuning attacks. We systematically discuss and characterize the risk of structural capability transfer in feed-forward Gaussian generators and summarize how explicit attribute interfaces, strong multi-view supervision, and low-cost adaptation amplify this threat.
    
    \item We propose GaussLock, a lightweight parameter-space immunization paradigm for feed-forward Gaussian generators, aiming to reduce the usability of unauthorized derivatives while preserving original source-task utility.
    
    \item We implement defense and source-utility preservation in the Gaussian parameter space by leveraging attribute-level outputs. This approach maintains viewpoint-agnostic behavior and avoids additional training burdens from rendering-loop dependencies.
    
    \item Extensive experiments demonstrate that under common downstream adaptation settings, GaussLock significantly reduces the target-domain usability of unauthorized derivatives while maintaining generation quality on the source tasks with favorable efficiency overhead.
\end{itemize}

\section{Related Work}
\label{sec:related}

\subsection{3D Generative Models}
Recent progress in 3D content generation has evolved significantly, shifting from implicit neural representations to explicit geometric primitives. Modern explicit 3D generative models generally fall into two main paradigms: point cloud generators and 3D Gaussian Splatting generators. Early methods often relied on point clouds~\cite{nichol2022pointegenerating3dpoint,chen2025harnessing,zhou2025recurrent,bastico2025rethinkingmetricsdiffusionarchitecture,liu2023robust,yao2025adversarial,zheng2023cgc,zhang2024jointly,zheng2026towards} to synthesize 3D shapes by predicting discrete spatial coordinates. While point clouds are straightforward, they often struggle to represent continuous surfaces and rich textures, frequently resulting in sparse or noisy visual outputs. In contrast, 3D Gaussian Splatting~\cite{kerbl20233d} has emerged as a superior and highly popular alternative. It defines a 3D object through continuous and explicit physical parameters including position, rotation, scale, opacity, and color. This paradigm offers high-fidelity reconstruction, photorealistic texture modeling, and real-time rendering capabilities. Consequently, state-of-the-art large-scale 3D generators such as LGM~\cite{tang2024lgm} and GaussianDreamer~\cite{yi2024gaussiandreamer} now leverage these Gaussian primitives to synthesize complex assets rapidly from diverse inputs like text or single images. We specifically select 3D Gaussian Splatting as the foundational representation for our defense framework because its highly explicit nature creates a significant security vulnerability. The transparent parameter space allows adversaries to easily manipulate fundamental physical attributes through gradient-based optimization during unauthorized fine-tuning. This direct path for geometric and structural theft makes Gaussian-based generators the most critical and ideal testbed for demonstrating
  our parameter-level immunization strategy under realistic attack settings.

\subsection{Model Immunization}
Securing intellectual property in deep learning is a critical challenge~\cite{liu2020dlgan}. While early protection methods utilized digital watermarking or model fingerprinting~\cite{uchida2017embedding, adi2018turning, lukas2019deep,wu2025robust,you2024two,zhong2020automated,luo2025diffw,sun2026grasp,zhang2026ghs,cao2026language} to identify unauthorized weight redistribution, recent focus has shifted toward proactive model immunization~\cite{shan2023glaze, shan2024nightshade, shan2020fawkes, wan2023poisoning, zheng2023targeted}. Immunization aims to render a model unusable for specific unauthorized tasks like fine-tuning or domain adaptation. In the 2D image ~\cite{sha2022fine,wang2021non,deng2024sophonnonfinetunablelearningrestrain,hong2025toward,hong2024improving,wang2024say,wang2026towards,peng2026dynamic} and language domains~\cite{rosati2024representationnoisingdefencemechanism,chen2025sddselfdegradeddefensemalicious,huang2024harmful,hong2024improving,hu2025adaptive,huangbooster,huang2024vaccine,qi2023fine,qisafety}, researchers have explored adversarial perturbations and dormant triggers to degrade performance when unauthorized data is used for training. These specific weight perturbations remain inactive during normal inference~\cite{liu2019soft, chen2019learning}. However, these 2D-centric defenses are insufficient for 3D tasks. 3D assets must maintain strict geometric consistency and physical plausibility across multiple viewpoints, and traditional pixel-level interventions cannot effectively govern the underlying physical attributes defining 3D structures. Furthermore, the protection of 3D generative models against fine-tuning attacks remains largely unexplored~\cite{zhao2026intellectual,wang2025machine,hong2025adlift,zhao2025rdsplat,huang2024gaussianmarker,yao2025adversarial}, even though differentiable rendering provides a direct path for structural theft. GaussLock addresses this critical gap as the first attempt to provide immunization for 3D generators. Instead of relying on 2D visual artifacts, we embed protection directly within the 3D parameter space. We utilize an attribute-aware strategy that triggers structural collapse during fine-tuning attacks.

\section{Methodology}
\label{sec:methods}

\begin{figure*}[!t]
\centering
\includegraphics[width=\textwidth]{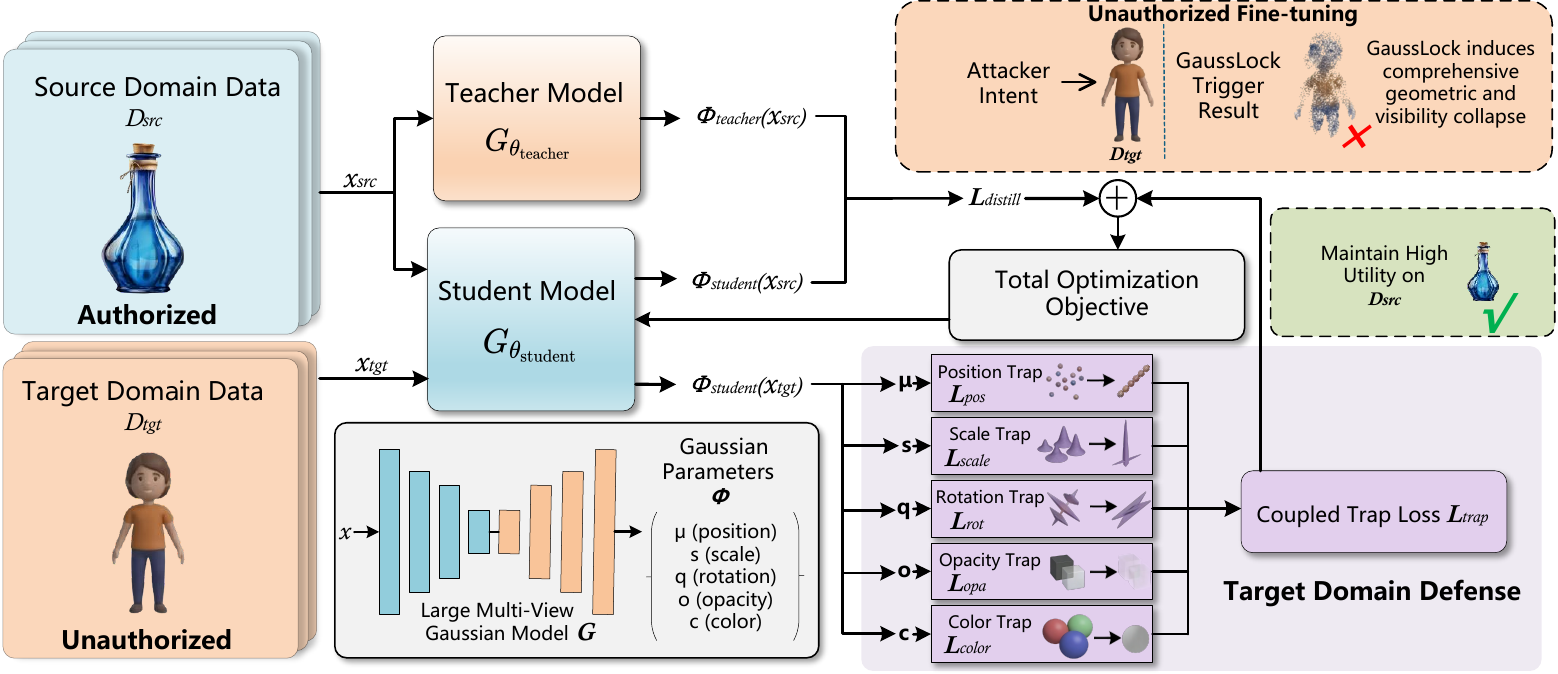}
\caption{Overview of the proposed GaussLock framework. Teacher-student distillation ensures high utility on original source tasks (checked green box). For the target domain, input views trigger a dormant parameter-space defense during unauthorized fine-tuning. Attribute-aware trap losses applied to explicit Gaussian parameters are averaged to form the Coupled Trap Loss ($\mathcal{L}_{trap}$), inducing comprehensive geometric and visibility collapse (crossed red box) to deny visual utility to adversaries while preserving source task fidelity.}
\label{fig:GaussLock_architecture}
\end{figure*}

\subsection{Overview}
This section introduces GaussLock, a parameter-space immunization framework designed to protect 3D generative models from unauthorized fine-tuning. Fig.~\ref{fig:GaussLock_architecture} illustrates the overall pipeline of GaussLock. First, in Section \ref{subsec:preliminaries}, we establish the preliminaries by detailing the Large Multi-View Gaussian Model (LGM) architecture and justifying its selection as our foundation. Second, in Section \ref{subsec:traps}, we elaborate on the core mechanism of GaussLock, which introduces five multi-attribute defensive traps targeting the explicit parameters of 3D Gaussians. Finally, in Section \ref{subsec:optimization}, we describe the training and optimization strategy, highlighting how source distillation is optimized to preserve generation quality on $\mathcal{D}_{src}$ while the coupled traps ensure robust defense on $\mathcal{D}_{tgt}$.

\subsection{Preliminaries}
\label{subsec:preliminaries}
Our defense framework is built upon the Large Multi-View Gaussian Model (LGM)~\cite{tang2024lgm}, a robust architecture for high-resolution 3D content generation. We select LGM as our foundational model because of its state-of-the-art feed-forward generation pipeline, which directly predicts explicit 3D Gaussian parameters from multi-view images using an asymmetric U-Net~\cite{ronneberger2015u}. This direct prediction mechanism bypasses the need for computationally expensive per-scene optimization, providing a highly efficient and transparent parameter space that is ideal for implementing and evaluating our proposed defense. Given a set of input views $I \in \mathbb{R}^{V \times H \times W \times 3}$, where $V$, $H$, and $W$ represent the number of views, height, and width respectively, the generator $\mathcal{G}_\theta$ parameterized by weights $\theta$ outputs the explicit physical parameters of the Gaussians: $\Phi = \mathcal{G}_\theta(I)$. These parameters are defined as $\Phi = \{\mu, o, s, q, c\}$, where $\mu \in \mathbb{R}^3$ is the position center, $o \in [0, 1]$ is the opacity, $s \in \mathbb{R}^3$ is the scaling factor, $q \in \mathbb{R}^4$ represents the rotation quaternion, and $c \in \mathbb{R}^3$ is the color of each primitive. In our security context, the pre-trained weights $\theta$ serve as the primary intellectual property. We consider a threat model where an adversary attempts to perform unauthorized fine-tuning on $\theta$ to extract proprietary capabilities. To comprehensively demonstrate defensive robustness, we evaluate against both prevalent Low-Rank Adaptation (LoRA)~\cite{hu2022lora} and full-parameter fine-tuning. Under the primary LoRA assumption, the adversary introduces low-rank update matrices $\Delta \theta = AB^\top$, where $A \in \mathbb{R}^{d \times r}$ and $B \in \mathbb{R}^{k \times r}$ are trainable matrices with rank $r \ll \min(d, k)$. The objective of GaussLock is to embed a dormant defense mechanism directly into this parameter space to disrupt these unauthorized adaptations.

\subsection{Multi-Attribute Defensive Traps}
\label{subsec:traps}
To counter the unauthorized parameter adaptations described previously, our defense strategy directly targets the explicit Gaussian attributes. We embed dormant objectives that force these physical parameters into degenerate states upon detecting unauthorized fine-tuning. Specifically, the core of GaussLock consists of five attribute-specific traps designed to systematically destroy the structural integrity of the 3D representation. These defensive objectives include the position trap, scale trap, rotation trap, color trap, and opacity trap. For the spatial position $\mu_i \in \mathbb{R}^3$, we compute the centralized covariance matrix $C_\mu$ as:
\begin{equation} C_\mu = \frac{1}{N}\sum_{i=1}^{N}(\mu_i - \bar{\mu})(\mu_i - \bar{\mu})^\top ,\label{eq:cov_p} \end{equation}
where $\bar{\mu}$ is the mean position. Then we formulate the position trap as:
\begin{equation} \mathcal{L}_{pos} = -\log\frac{\lambda_{\max}(C_\mu)}{\lambda_{\min}(C_\mu)+\epsilon} ,\label{eq:pos} \end{equation}
where $\lambda_{\max}(C)$ and $\lambda_{\min}(C)$ denote the maximum and minimum eigenvalues of a covariance or scatter matrix $C$, and $\epsilon > 0$ is a numerical stability term. This trap effectively collapses the normal 3D spatial distribution into degraded low-dimensional structures like lines or planes to ruin volumetric consistency. Similarly, the scale trap targets the disparity among the squared scaling components $u_i = s_i^2$ for each Gaussian by minimizing:
\begin{equation} \mathcal{L}_{scale} = -\frac{1}{N}\sum_{i=1}^{N}\log\frac{\max(u_i)}{\min(u_i)+\epsilon} ,\label{eq:scale} \end{equation}
which forces the primitives into extremely unstable needle-like or flake-like shapes that destroy geometric support. To disrupt directional diversity and local surface normals, we extract the local main axis for each unit quaternion $q_i \in \mathbb{R}^4$ as $r_i = R(q_i)e_z$, where $R(q_i)$ is the rotation matrix and $e_z = [0, 0, 1]^\top$, construct the structure tensor:
\begin{equation} T_r = \frac{1}{N}\sum_{i=1}^N r_i r_i^\top ,\label{eq:tensor_r} \end{equation}
and apply the rotation trap to force all Gaussian axes to align:
\begin{equation} \mathcal{L}_{rot} = -\log\frac{\lambda_{\max}(T_r)}{\lambda_{\min}(T_r)+\epsilon} .\label{eq:rot} \end{equation}
The color trap follows an identical anisotropic logic using the centralized color covariance matrix:
\begin{equation} C_c = \frac{1}{N}\sum_{i=1}^N(c_i-\bar{c})(c_i-\bar{c})^\top ,\label{eq:cov_c} \end{equation}
based on the RGB color $c_i \in \mathbb{R}^3$ and the mean color $\bar{c}$, compressing the color distribution to eradicate rich textures and material variations:
\begin{equation} \mathcal{L}_{color} = -\log\frac{\lambda_{\max}(C_c)}{\lambda_{\min}(C_c)+\epsilon} .\label{eq:color} \end{equation}
The opacity trap targets the visibility support of the Gaussian representation by suppressing the high-value tail that dominates foreground rendering. Since scenes naturally contain numerous low-opacity background Gaussians, applying a uniform penalty would cause the objective function to be dominated by these nearly transparent elements. This would critically weaken the suppression of visible foreground structures. Therefore, we exclusively retain the top-$k$ elements with the highest opacity for optimization. Furthermore, let the opacity be defined as $o = \sigma(z)$, where $z$ represents the underlying logit. Directly penalizing $\log(o)$ yields a gradient of $1 - o$ with respect to $z$. As $o$ approaches 1, this gradient diminishes, making the most visible Gaussians harder to continuously suppress due to gradient saturation. To resolve this, we instead optimize the smoothed logits defined as:
\begin{equation} \ell_i = \operatorname{logit}(o_i) \approx \log(o_i + \epsilon) - \log(1 - o_i + \epsilon) .\label{eq:logit} \end{equation}
We then formulate the opacity trap as:
\begin{equation} \mathcal{L}_{opa} = \frac{1}{k}\sum\limits_{j=1}^{k}\ell_{(j)} .\label{eq:opa} \end{equation}
Minimizing this objective consistently suppresses the most visually contributing primitives and drives the representation into a degraded state of visibility collapse. Finally, to ensure comprehensive protection across all physical attributes, we aggregate the $m$ enabled traps $\{\mathcal{L}_1, \dots, \mathcal{L}_m\}$ by calculating their average, computed as:
\begin{equation} \mathcal{L}_{trap} = \frac{1}{m} \sum\limits_{j=1}^{m} \mathcal{L}_j .\label{eq:coupled_trap} \end{equation}
This combined objective guarantees that any unauthorized adaptation incurs a simultaneous and comprehensive penalty across the entire parameter space.

\subsection{Training and Optimization}
\label{subsec:optimization}
The final objective balances model immunization with utility preservation through a highly efficient optimization strategy. We employ parameter-space source distillation to anchor the performance on $\mathcal{D}_{src}$, ensuring that the original generative capabilities are strictly maintained. We designate the original frozen model as the teacher and the defended model as the student. The distillation loss is defined as:
\begin{equation} \mathcal{L}_{distill} = \frac{1}{N}\left\|\Phi_{student}(x_{src})-\Phi_{teacher}(x_{src})\right\|_1 ,\label{eq:distill} \end{equation}
where $x_{src} \in \mathcal{D}_{src}$. Crucially, because this distillation occurs entirely within the Gaussian parameter space, the teacher outputs can be pre-computed offline and cached, completely eliminating the need for repeated forward passes and rendering loops during training. While distillation preserves the knowledge on $\mathcal{D}_{src}$, the trap losses are actively optimized to construct a defense against unauthorized adaptation on $\mathcal{D}_{tgt}$. The total objective evaluates the coupled traps directly on the model weights $\theta$, formulated as:
\begin{equation} \mathcal{L}_{def} = \lambda_{distill}\mathcal{L}_{distill} + \lambda_{trap}\mathcal{L}_{trap} ,\label{eq:total_loss} \end{equation}
where $\lambda_{distill}$ and $\lambda_{trap}$ are weighting coefficients. The complete training iteration involves sequentially computing the distillation loss on a source batch using the offline cache, calculating the trap loss on a target batch, and executing a single-step first-order backward pass to update the weights. This unified approach effectively locks the pre-trained knowledge, ensuring high-fidelity generation on $\mathcal{D}_{src}$ while guaranteeing structural collapse upon fine-tuning on $\mathcal{D}_{tgt}$.

\section{Experiments}
\label{sec:Experiments}

\subsection{Experimental Settings}

\subsubsection{Benchmarks and Baselines}
\label{subsec:benchmarks}

To evaluate the effectiveness of GaussLock, we focus on five representative categories from our target datasets including shoe, plant, dish, bowl, and box. We evaluate the defense using four primary metrics including Target PSNR~\cite{hore2010image}, Target LPIPS~\cite{zhang2018unreasonable}, Source PSNR, and Source LPIPS. The target metrics assess the quality of unauthorized reconstructions on the target domain, where a successful defense is characterized by a significant decrease in PSNR and an increase in LPIPS. Conversely, the source metrics measure the generative fidelity on the authorized domain, with high PSNR and low LPIPS indicating that the original capabilities of the model are effectively preserved. We compare our method against two baselines including an unprotected baseline involving fine-tuning without any defensive measures and a
  naive-unlearning approach that optimizes a combination of a reversed task loss and a source distillation loss under matched training
  settings.

To maintain the foundational generative performance, we use Objaverse, the original training dataset of the Large Multi-View Gaussian Model, as the authorized source domain. We define the sources of defense and attack data as two distinct domains to formulate two primary attack scenarios. The first is an ideal scenario where defense and attack data originate from the same distribution. The second is a robust scenario characterized by a distribution shift between the defense and attack data. We adopt a directional notation to represent these configurations. For example, a robust scenario utilizing Google Scanned Objects~\cite{downs2022google} for defense and OmniObject3D~\cite{wu2023omniobject3d} for the attack is denoted as GSO $\rightarrow$ Omni. In the ideal Omni $\rightarrow$ Omni setting, we randomly select 40\% of the objects from OmniObject3D to construct the defense and reserve the remaining 60\% to simulate the unauthorized attack. For the robust evaluations in both GSO $\rightarrow$ Omni and Omni $\rightarrow$ GSO, we utilize all available objects from the respective datasets. Since the Google Scanned Objects dataset only provides descriptive tags without explicit category labels, we filter and match these tags to precisely align with the established categories in OmniObject3D. For every object in our experiments, we sample five sets of orthogonal views from different angles to construct the training data.

\subsubsection{Implementation Details}
\label{subsec:implementation}

During the optimization process, both the defense and attack phases employ a batch size of 4 and utilize the AdamW optimizer with a learning rate of 3e-5, alongside a weight decay of 0.05 and a gradient clipping of 1.0. Unless specified otherwise, the default configuration trains the defense mechanism for 100 steps and runs the fine-tuning attack for 400 steps. For LoRA attacks, we inject trainable modules into the MVAttention layers of the U-Net architecture. We set both the rank and the scaling factor alpha to 16 alongside a dropout rate of 0.1. Besides this primary evaluation, we also conduct full-parameter fine-tuning attack experiments to comprehensively verify the robustness of our framework. To compute the losses, we alternate source and target batches at the batch level, using 80\% source batches for the distillation loss and 20\% target batches
  for the trap loss. For our GaussLock method, we set the distillation loss weight $\lambda_{distill} = 400$ and the trap loss weight $\lambda_{trap} = 20$. To effectively benchmark our approach, we compare it against a naive-unlearning baseline. We carefully calibrate the loss weights of this baseline to guarantee comparable source domain fidelity and ensure a fair evaluation. Further implementation details regarding this baseline are provided in the \textcolor{red}{Supplementary Material}. All five attribute-space traps are activated simultaneously during training, and to prevent large amounts of transparent background tokens from diluting the optimization signal, we exclusively calculate the opacity trap loss on the top 2\% of Gaussian points with the highest initial opacity values at
  initialization.

\subsection{Main Results}
\label{subsec:main_results}

For intra-domain attacks on the Omni dataset, as shown in Table \ref{tab:omni_omni_lora} and Table \ref{tab:omni_omni_full}, GaussLock triggers a rapid structural degradation. While Table \ref{tab:omni_omni_lora} focuses on parameter-efficient methods, Table \ref{tab:omni_omni_full} further demonstrates our defense's resilience against more aggressive full-parameter fine-tuning attacks. For most categories, GaussLock forces the model into an opacity mode collapse, locking target metrics at persistently low levels right from the beginning of the attack.

The defensive capability of our framework extends robustly to cross-domain scenarios where the attacker fine-tunes the model across different dataset distributions. As illustrated in Table \ref{tab:gso_omni_lora}, our attribute-aware traps continue to disrupt the optimization process effectively despite severe domain shifts. We also provide the corresponding results for the Omni to GSO configuration in the \textcolor{red}{Supplementary Material}. GaussLock restricts the target PSNR to stagnant levels across all attack steps to prevent the attacker from extracting structural information. These findings collectively confirm that the embedded attribute-space traps are dataset-agnostic. They provide a robust barrier that actively neutralizes multiple fine-tuning methods while remaining consistently effective against parameter-efficient attacks like LoRA. We note that, for a subset of categories across multiple evaluation settings, once the defended model collapses to a fully transparent
  rendering state, the composited output becomes fixed under our rendering setup; consequently, the corresponding PSNR and LPIPS remain
  unchanged across later attack steps.

\begin{table*}[t]
\centering
\caption{Quantitative results of intra-domain LoRA fine-tuning attacks (Omni $\rightarrow$ Omni). Target metrics are reported across different optimization steps.}
\label{tab:omni_omni_lora}
\resizebox{\textwidth}{!}{
\begin{tabular}{l|l|cc|cc|cc|cc|cc}
\toprule
\multirow{2}{*}{\textbf{Category}} & \multirow{2}{*}{\textbf{Method}} & \multicolumn{2}{c|}{\textbf{Source Utility (Avg.)}} & \multicolumn{2}{c|}{\textbf{Target (100 Steps)}} & \multicolumn{2}{c|}{\textbf{Target (200 Steps)}} & \multicolumn{2}{c|}{\textbf{Target (300 Steps)}} & \multicolumn{2}{c}{\textbf{Target (400 Steps)}} \\
\cmidrule(lr){3-4} \cmidrule(lr){5-6} \cmidrule(lr){7-8} \cmidrule(lr){9-10} \cmidrule(lr){11-12}
& & PSNR $\uparrow$ & LPIPS $\downarrow$ & PSNR $\downarrow$ & LPIPS $\uparrow$ & PSNR $\downarrow$ & LPIPS $\uparrow$ & PSNR $\downarrow$ & LPIPS $\uparrow$ & PSNR $\downarrow$ & LPIPS $\uparrow$ \\
\midrule
\multirow{3}{*}{shoe} & baseline & 22.52 & 0.0722 & 23.17 & 0.0573 & 23.76 & 0.0527 & 24.35 & 0.0490 & 24.85 & 0.0464 \\
& naive-unlearning & 21.80 & 0.0822 & 16.46 & 0.1517 & 19.57 & 0.1150 & 21.30 & 0.0854 & 23.11 & 0.0668 \\
& ours (GaussLock) & 21.25 & 0.0838 & \textbf{13.96} & \textbf{0.1920} & \textbf{13.96} & \textbf{0.1920} & \textbf{13.96} & \textbf{0.1920} & \textbf{13.96} & \textbf{0.1920} \\
\midrule
\multirow{3}{*}{plant} & baseline & 22.52 & 0.0722 & 22.61 & 0.0740 & 23.22 & 0.0679 & 23.78 & 0.0630 & 24.22 & 0.0600 \\
& naive-unlearning & 21.72 & 0.0855 & \textbf{11.92} & \textbf{0.2251} & 13.58 & 0.1894 & 19.43 & 0.1254 & 21.32 & 0.1028 \\
& ours (GaussLock) & 22.21 & 0.0749 & 12.14 & 0.2105 & \textbf{12.26} & \textbf{0.2091} & \textbf{12.35} & \textbf{0.2082} & \textbf{12.43} & \textbf{0.2071} \\
\midrule
\multirow{3}{*}{dish} & baseline & 22.52 & 0.0722 & 23.92 & 0.0694 & 24.68 & 0.0616 & 25.25 & 0.0567 & 25.69 & 0.0533 \\
& naive-unlearning & 21.96 & 0.0780 & 16.59 & 0.1626 & 20.17 & 0.1234 & 21.79 & 0.0986 & 22.98 & 0.0799 \\
& ours (GaussLock) & 21.94 & 0.0768 & \textbf{12.21} & \textbf{0.2372} & \textbf{12.21} & \textbf{0.2372} & \textbf{12.21} & \textbf{0.2372} & \textbf{12.21} & \textbf{0.2372} \\
\midrule
\multirow{3}{*}{bowl} & baseline & 22.52 & 0.0722 & 21.34 & 0.0985 & 22.03 & 0.0862 & 22.61 & 0.0804 & 22.93 & 0.0774 \\
& naive-unlearning & 22.18 & 0.0791 & 11.29 & \textbf{0.3173} & \textbf{11.28} & \textbf{0.2982} & \textbf{11.28} & 0.2977 & 11.29 & 0.2974 \\
& ours (GaussLock) & 22.17 & 0.0754 & \textbf{11.28} & 0.2981 & \textbf{11.28} & 0.2981 & \textbf{11.28} & \textbf{0.2981} & \textbf{11.28} & \textbf{0.2981} \\
\midrule
\multirow{3}{*}{box} & baseline & 22.52 & 0.0722 & 23.88 & 0.0803 & 24.34 & 0.0736 & 24.74 & 0.0679 & 25.06 & 0.0641 \\
& naive-unlearning & 21.33 & 0.0861 & 18.60 & 0.1768 & 19.59 & 0.1548 & 20.99 & 0.1313 & 22.13 & 0.1104 \\
& ours (GaussLock) & 19.94 & 0.1010 & \textbf{11.37} & \textbf{0.2878} & \textbf{11.37} & \textbf{0.2878} & \textbf{11.37} & \textbf{0.2878} & \textbf{11.37} & \textbf{0.2878} \\
\bottomrule
\end{tabular}
}
\end{table*}

\begin{table*}[t]
\centering
\caption{Quantitative results of cross-domain LoRA fine-tuning attacks (GSO $\rightarrow$ Omni).}
\label{tab:gso_omni_lora}
\resizebox{\textwidth}{!}{
\begin{tabular}{l|l|cc|cc|cc|cc|cc}
\toprule
\multirow{2}{*}{\textbf{Category}} & \multirow{2}{*}{\textbf{Method}} & \multicolumn{2}{c|}{\textbf{Source Utility (Avg.)}} & \multicolumn{2}{c|}{\textbf{Target (100 Steps)}} & \multicolumn{2}{c|}{\textbf{Target (200 Steps)}} & \multicolumn{2}{c|}{\textbf{Target (300 Steps)}} & \multicolumn{2}{c}{\textbf{Target (400 Steps)}} \\
\cmidrule(lr){3-4} \cmidrule(lr){5-6} \cmidrule(lr){7-8} \cmidrule(lr){9-10} \cmidrule(lr){11-12}
& & PSNR $\uparrow$ & LPIPS $\downarrow$ & PSNR $\downarrow$ & LPIPS $\uparrow$ & PSNR $\downarrow$ & LPIPS $\uparrow$ & PSNR $\downarrow$ & LPIPS $\uparrow$ & PSNR $\downarrow$ & LPIPS $\uparrow$ \\
\midrule
\multirow{3}{*}{shoe} & baseline & 22.52 & 0.0722 & 23.33 & 0.0563 & 23.96 & 0.0515 & 24.57 & 0.0477 & 25.07 & 0.0451 \\
& naive-unlearning & 21.40 & 0.0812 & 19.49 & 0.0913 & 21.66 & 0.0698 & 22.87 & 0.0596 & 23.65 & 0.0535 \\
& ours (GaussLock) & 20.67 & 0.0890 & \textbf{13.71} & \textbf{0.1916} & \textbf{13.71} & \textbf{0.1915} & \textbf{13.73} & \textbf{0.1910} & \textbf{13.76} & \textbf{0.1900} \\
\midrule
\multirow{3}{*}{plant} & baseline & 22.52 & 0.0722 & 22.64 & 0.0734 & 23.27 & 0.0674 & 23.81 & 0.0627 & 24.30 & 0.0593 \\
& naive-unlearning & 21.72 & 0.0923 & 20.72 & 0.1151 & 21.55 & 0.0928 & 22.28 & 0.0777 & 23.09 & 0.0695 \\
& ours (GaussLock) & 21.52 & 0.0801 & \textbf{16.15} & \textbf{0.1572} & \textbf{20.69} & \textbf{0.1032} & \textbf{21.43} & \textbf{0.0901} & \textbf{22.26} & \textbf{0.0825} \\
\midrule
\multirow{3}{*}{dish} & baseline & 22.52 & 0.0722 & 24.21 & 0.0666 & 24.93 & 0.0594 & 25.56 & 0.0543 & 26.01 & 0.0511 \\
& naive-unlearning & 21.46 & 0.0848 & 18.43 & 0.1492 & 21.15 & 0.1081 & 23.02 & 0.0799 & 24.06 & 0.0659 \\
& ours (GaussLock) & 20.57 & 0.0935 & \textbf{12.20} & \textbf{0.2400} & \textbf{12.20} & \textbf{0.2400} & \textbf{12.20} & \textbf{0.2400} & \textbf{12.20} & \textbf{0.2400} \\
\midrule
\multirow{3}{*}{bowl} & baseline & 22.52 & 0.0722 & 21.52 & 0.0965 & 22.22 & 0.0839 & 22.80 & 0.0779 & 23.18 & 0.0745 \\
& naive-unlearning & 20.69 & 0.0924 & 14.71 & 0.2633 & 17.37 & 0.1892 & 18.76 & 0.1473 & 19.88 & 0.1212 \\
& ours (GaussLock) & 21.49 & 0.0804 & \textbf{11.28} & \textbf{0.2909} & \textbf{11.28} & \textbf{0.2909} & \textbf{11.28} & \textbf{0.2909} & \textbf{11.28} & \textbf{0.2909} \\
\midrule
\multirow{3}{*}{box} & baseline & 22.52 & 0.0722 & 23.96 & 0.0777 & 24.51 & 0.0702 & 25.03 & 0.0645 & 25.46 & 0.0605 \\
& naive-unlearning & 20.53 & 0.0907 & 16.60 & 0.1998 & 19.35 & 0.1366 & 21.38 & 0.1095 & 22.36 & 0.0946 \\
& ours (GaussLock) & 20.96 & 0.0891 & \textbf{13.35} & \textbf{0.2466} & \textbf{15.47} & \textbf{0.2172} & \textbf{19.34} & \textbf{0.1615} & \textbf{21.89} & \textbf{0.1216} \\
\bottomrule
\end{tabular}
}
\end{table*}

\begin{table*}[t]
\centering
\caption{Quantitative results of intra-domain full-parameter fine-tuning attacks (Omni $\rightarrow$ Omni).}
\label{tab:omni_omni_full}
\resizebox{\textwidth}{!}{
\begin{tabular}{l|l|cc|cc|cc|cc|cc}
\toprule
\multirow{2}{*}{\textbf{Category}} & \multirow{2}{*}{\textbf{Method}} & \multicolumn{2}{c|}{\textbf{Source Utility (Avg.)}} & \multicolumn{2}{c|}{\textbf{Target (100 Steps)}} & \multicolumn{2}{c|}{\textbf{Target (200 Steps)}} & \multicolumn{2}{c|}{\textbf{Target (300 Steps)}} & \multicolumn{2}{c}{\textbf{Target (400 Steps)}} \\
\cmidrule(lr){3-4} \cmidrule(lr){5-6} \cmidrule(lr){7-8} \cmidrule(lr){9-10} \cmidrule(lr){11-12}
& & PSNR $\uparrow$ & LPIPS $\downarrow$ & PSNR $\downarrow$ & LPIPS $\uparrow$ & PSNR $\downarrow$ & LPIPS $\uparrow$ & PSNR $\downarrow$ & LPIPS $\uparrow$ & PSNR $\downarrow$ & LPIPS $\uparrow$ \\
\midrule
\multirow{3}{*}{shoe} & baseline & 22.52 & 0.0722 & 26.11 & 0.0354 & 26.35 & 0.0336 & 26.55 & 0.0320 & 26.54 & 0.0314 \\
& naive-unlearning & 21.80 & 0.0822 & 26.24 & 0.0356 & 26.43 & 0.0338 & 26.46 & 0.0329 & 26.79 & 0.0317 \\
& ours (GaussLock) & 21.25 & 0.0838 & \textbf{13.96} & \textbf{0.1920} & \textbf{13.96} & \textbf{0.1920} & \textbf{13.96} & \textbf{0.1920} & \textbf{13.96} & \textbf{0.1920} \\
\midrule
\multirow{3}{*}{plant} & baseline & 22.52 & 0.0722 & 25.88 & 0.0459 & 26.14 & 0.0416 & 26.52 & 0.0383 & 26.46 & 0.0375 \\
& naive-unlearning & 21.72 & 0.0855 & 25.90 & 0.0458 & 26.40 & 0.0416 & 26.69 & 0.0389 & 26.82 & 0.0366 \\
& ours (GaussLock) & 22.21 & 0.0749 & \textbf{25.59} & \textbf{0.0483} & \textbf{26.25} & \textbf{0.0421} & \textbf{26.30} & \textbf{0.0396} & \textbf{26.35} & \textbf{0.0391} \\
\midrule
\multirow{3}{*}{dish} & baseline & 22.52 & 0.0722 & 26.76 & 0.0403 & 27.98 & 0.0355 & 22.40 & 0.0578 & 24.31 & 0.0389 \\
& naive-unlearning & 21.96 & 0.0780 & 27.36 & 0.0398 & 27.76 & 0.0379 & 28.20 & 0.0346 & 22.47 & 0.0462 \\
& ours (GaussLock) & 21.94 & 0.0768 & \textbf{12.21} & \textbf{0.2372} & \textbf{12.21} & \textbf{0.2372} & \textbf{12.21} & \textbf{0.2372} & \textbf{12.21} & \textbf{0.2372} \\
\midrule
\multirow{3}{*}{bowl} & baseline & 22.52 & 0.0722 & 24.53 & 0.0596 & 24.87 & 0.0566 & 25.31 & 0.0531 & 25.20 & 0.0522 \\
& naive-unlearning & 22.18 & 0.0791 & 24.44 & 0.0622 & 25.01 & 0.0571 & 25.09 & 0.0534 & 25.44 & 0.0518 \\
& ours (GaussLock) & 22.17 & 0.0754 & \textbf{11.28} & \textbf{0.2981} & \textbf{11.28} & \textbf{0.2981} & \textbf{11.28} & \textbf{0.2981} & \textbf{11.28} & \textbf{0.2981} \\
\midrule
\multirow{3}{*}{box} & baseline & 22.52 & 0.0722 & 26.24 & 0.0510 & 26.39 & 0.0497 & 26.57 & 0.0485 & 26.15 & 0.0503 \\
& naive-unlearning & 21.33 & 0.0861 & 25.81 & 0.0529 & 26.22 & 0.0496 & 25.97 & 0.0520 & 25.37 & 0.0535 \\
& ours (GaussLock) & 19.94 & 0.1010 & \textbf{11.37} & \textbf{0.2878} & \textbf{11.37} & \textbf{0.2878} & \textbf{11.37} & \textbf{0.2878} & \textbf{11.37} & \textbf{0.2878} \\
\bottomrule
\end{tabular}
}
\end{table*}

\subsection{Analysis}
\label{subsec:ablation}

To comprehensively evaluate the mechanisms and robustness of GaussLock, we conduct a series of detailed analyses including extensive ablation studies. All subsequent experiments exclusively employ LoRA for the fine-tuning attacks. Please note that the \textbf{Generalization Across 3D Generative Models}, \textbf{Hyper-Parameter Ablation Analysis} and \textbf{Efficiency Analysis} are provided in the \textcolor{red}{Supplementary Material} for further reference.


\subsubsection{Impact of Model Initialization}
\label{subsubsec:initialization}

To demonstrate the necessity of protecting pre-trained intellectual property, we compare an unprotected pre-trained model against a randomly initialized model under identical fine-tuning steps. As shown in Table \ref{tab:ablation_init}, exploiting pre-trained weights provides a significant advantage. Unlike a randomly initialized model that fails to capture meaningful geometric information within a limited training budget, the pre-trained model allows attackers to quickly achieve superior reconstructions with high visual fidelity. This confirms that malicious adaptation relies heavily on pre-trained knowledge, underscoring the critical need for GaussLock to prevent adversaries from freely extracting high-value structural priors.

\begin{table}[htbp]
\centering
 
\caption{Ablation study on model initialization (GSO $\rightarrow$ Omni). We compare the target domain reconstruction quality (after 400 attack steps) between a model initialized with pre-trained weights and one with random weights.}

\label{tab:ablation_init}

\resizebox{\columnwidth}{!}{
\begin{tabular}{l|l|c|c}
\toprule
\textbf{Category} & \textbf{Initialization} & \textbf{Target PSNR} $\uparrow$ & \textbf{Target LPIPS} $\downarrow$ \\
\midrule
\multirow{2}{*}{shoe} & random    & 7.80 & 0.6159 \\
                      & pre-train & \textbf{25.03} & \textbf{0.0452} \\
\midrule
\multirow{2}{*}{plant} & random    & 7.67 & 0.6037 \\
                       & pre-train & \textbf{24.28} & \textbf{0.0595} \\
\midrule
\multirow{2}{*}{dish} & random    & 8.04 & 0.6213 \\
                      & pre-train & \textbf{26.03} & \textbf{0.0511} \\
\midrule
\multirow{2}{*}{bowl} & random    & 8.27 & 0.6193 \\
                      & pre-train & \textbf{23.20} & \textbf{0.0743} \\
\midrule
\multirow{2}{*}{box} & random    & 7.90 & 0.6346 \\
                     & pre-train & \textbf{25.44} & \textbf{0.0607} \\
\bottomrule
\end{tabular}
}
\end{table}

\subsubsection{Robustness against Diverse Fine-tuning Attacks}
\label{subsubsec:attack_robustness}

To comprehensively evaluate the durability of GaussLock, we simulate diverse attack configurations in a cross-domain setting denoted as GSO
  $\rightarrow$ Omni. Table~\ref{tab:ablation_opt} demonstrates that GaussLock effectively preserves the degraded state regardless of specific
  optimizers such as AdamW or SGD and varying learning rates. As shown in Table~\ref{tab:ablation_lora}, our defense also maintains consistent
  resistance across different LoRA ranks. Finally, evaluations under extended training budgets of 800 and 1600 steps in
  Table~\ref{tab:ablation_extended_budget} confirm the long-term persistence of our embedded traps. Although prolonged optimization eventually
  recovers minor structural fidelity, GaussLock consistently outperforms the naive-unlearning baseline across diverse attack configurations,
  demonstrating highly reliable and practically enduring robustness.


\begin{table}[htbp]
\centering
\caption{Ablation on different optimizers and learning rates (GSO $\rightarrow$ Omni). Target metrics are recorded at 200 and 400 steps. The Setting column indicates the specific optimizer paired with its corresponding learning rate.}
\label{tab:ablation_opt}
\resizebox{\columnwidth}{!}{
\begin{tabular}{l|l|cc|cc}
\toprule
\multirow{2}{*}{\textbf{Setting}} & \multirow{2}{*}{\textbf{Method}} & \multicolumn{2}{c|}{\textbf{200 Steps}} & \multicolumn{2}{c}{\textbf{400 Steps}} \\
\cmidrule(lr){3-4} \cmidrule(lr){5-6}
& & PSNR $\downarrow$ & LPIPS $\uparrow$ & PSNR $\downarrow$ & LPIPS $\uparrow$ \\
\midrule
\multirow{2}{*}{AdamW 3e-6}
& naive-unlearning & 19.13 & 0.1392 & 19.70 & 0.1326 \\
& ours & \textbf{12.60} & \textbf{0.1955} & \textbf{13.07} & \textbf{0.1896} \\
\midrule
\multirow{2}{*}{AdamW 3e-5}
& naive-unlearning & 21.55 & 0.0928 & 23.09 & 0.0695 \\
& ours & \textbf{20.69} & \textbf{0.1032} & \textbf{22.26} & \textbf{0.0825} \\
\midrule
\multirow{2}{*}{SGD 3e-5}
& naive-unlearning & 18.65 & 0.1438 & 18.66 & 0.1437 \\
& ours & \textbf{12.31} & \textbf{0.1992} & \textbf{12.32} & \textbf{0.1991} \\
\midrule
\multirow{2}{*}{SGD 3e-4}
& naive-unlearning & 18.76 & 0.1428 & 18.88 & 0.1418 \\
& ours & \textbf{12.37} & \textbf{0.1985} & \textbf{12.44} & \textbf{0.1977} \\
\midrule
\multirow{2}{*}{SGD 3e-3}
& naive-unlearning & 19.73 & 0.1325 & 20.48 & 0.1196 \\
& ours & \textbf{13.07} & \textbf{0.1901} & \textbf{19.43} & \textbf{0.1217} \\
\bottomrule
\end{tabular}
}
\end{table}

\begin{table}[htbp]
\centering
\caption{Ablation on LoRA ranks (GSO $\rightarrow$ Omni). Target metrics are recorded at 200 and 400 steps. The Setting column indicates the specific rank used for the adaptation.}
\label{tab:ablation_lora}
\resizebox{\columnwidth}{!}{
\begin{tabular}{l|l|cc|cc}
\toprule
\multirow{2}{*}{\textbf{Setting}} & \multirow{2}{*}{\textbf{Method}} & \multicolumn{2}{c|}{\textbf{200 Steps}} & \multicolumn{2}{c}{\textbf{400 Steps}} \\
\cmidrule(lr){3-4} \cmidrule(lr){5-6}
& & PSNR $\downarrow$ & LPIPS $\uparrow$ & PSNR $\downarrow$ & LPIPS $\uparrow$ \\
\midrule
\multirow{2}{*}{R8} 
& naive-unlearning & 21.04 & 0.1069 & 22.06 & 0.0801 \\
& ours & \textbf{18.81} & \textbf{0.1300} & \textbf{21.24} & \textbf{0.0920} \\
\midrule
\multirow{2}{*}{R16}
& naive-unlearning & 21.55 & 0.0928 & 23.09 & 0.0695 \\
& ours & \textbf{20.69} & \textbf{0.1032} & \textbf{22.26} & \textbf{0.0825} \\
\midrule
\multirow{2}{*}{R32} 
& naive-unlearning & 22.36 & 0.0774 & 24.10 & 0.0620 \\
& ours & \textbf{21.38} & \textbf{0.0911} & \textbf{23.52} & \textbf{0.0710} \\
\bottomrule
\end{tabular}
}
\end{table}

\begin{table}[htbp]
\centering
\caption{Ablation on extended training budgets (GSO $\rightarrow$ Omni). Target metrics are recorded at different intervals for total optimization budgets of 800 and 1600 steps.}
\label{tab:ablation_extended_budget}
\resizebox{\columnwidth}{!}{
\begin{tabular}{l|l|cc|cc|cc|cc}
\toprule
\multicolumn{10}{c}{\textbf{Extended Budget: 800 Steps}} \\
\midrule
\multirow{2}{*}{\textbf{Setting}} & \multirow{2}{*}{\textbf{Method}} & \multicolumn{2}{c|}{\textbf{200 Steps}} & \multicolumn{2}{c|}{\textbf{400 Steps}} & \multicolumn{2}{c|}{\textbf{600 Steps}} & \multicolumn{2}{c}{\textbf{800 Steps}} \\
\cmidrule(lr){3-4} \cmidrule(lr){5-6} \cmidrule(lr){7-8} \cmidrule(lr){9-10}
& & PSNR $\downarrow$ & LPIPS $\uparrow$ & PSNR $\downarrow$ & LPIPS $\uparrow$ & PSNR $\downarrow$ & LPIPS $\uparrow$ & PSNR $\downarrow$ & LPIPS $\uparrow$ \\
\midrule
\multirow{2}{*}{800 Steps}
& N.U. & 21.55 & 0.0919 & 23.10 & 0.0693 & 24.08 & 0.0620 & 24.68 & 0.0578 \\
& ours & \textbf{20.69} & \textbf{0.1027} & \textbf{22.26} & \textbf{0.0826} & \textbf{23.60} & \textbf{0.0698} & \textbf{24.15} & \textbf{0.0639} \\
\midrule
\midrule
\multicolumn{10}{c}{\textbf{Highly Aggressive Budget: 1600 Steps}} \\
\midrule
\multirow{2}{*}{\textbf{Setting}} & \multirow{2}{*}{\textbf{Method}} & \multicolumn{2}{c|}{\textbf{400 Steps}} & \multicolumn{2}{c|}{\textbf{800 Steps}} & \multicolumn{2}{c|}{\textbf{1200 Steps}} & \multicolumn{2}{c}{\textbf{1600 Steps}} \\
\cmidrule(lr){3-4} \cmidrule(lr){5-6} \cmidrule(lr){7-8} \cmidrule(lr){9-10}
& & PSNR $\downarrow$ & LPIPS $\uparrow$ & PSNR $\downarrow$ & LPIPS $\uparrow$ & PSNR $\downarrow$ & LPIPS $\uparrow$ & PSNR $\downarrow$ & LPIPS $\uparrow$ \\
\midrule
\multirow{2}{*}{1600 Steps}
& N.U. & 23.10 & 0.0693 & 24.69 & 0.0578 & 25.31 & 0.0534 & 25.62 & 0.0510 \\
& ours & \textbf{22.26} & \textbf{0.0826} & \textbf{24.19} & \textbf{0.0636} & \textbf{24.80} & \textbf{0.0580} & \textbf{25.15} & \textbf{0.0549} \\
\bottomrule
\end{tabular}
}
\end{table}

\begin{table}[h]
\centering
\caption{Ablation study on multi-category joint attacks (GSO $\rightarrow$ Omni). Target metrics represent the combined performance across all categories involved in the fine-tuning attack and are recorded at 200 and 400 steps.}
\label{tab:ablation_multi_cat}
\resizebox{\columnwidth}{!}{
\begin{tabular}{l|l|cc|cc|cc}
\toprule
\multirow{2}{*}{\textbf{Joint Categories}} & \multirow{2}{*}{\textbf{Method}} & \multicolumn{2}{c|}{\textbf{Source Utility Avg.}} & \multicolumn{2}{c|}{\textbf{Target 200 Steps}} & \multicolumn{2}{c}{\textbf{Target 400 Steps}} \\
\cmidrule(lr){3-4} \cmidrule(lr){5-6} \cmidrule(lr){7-8}
& & PSNR $\uparrow$ & LPIPS $\downarrow$ & PSNR $\downarrow$ & LPIPS $\uparrow$ & PSNR $\downarrow$ & LPIPS $\uparrow$ \\
\midrule
\multirow{3}{*}{bowl, shoe} & baseline & 22.52 & 0.0722 & 23.12 & 0.0663 & 24.05 & 0.0587 \\
& naive-unlearning & 21.35 & 0.0844 & 20.87 & 0.0940 & 22.61 & 0.0709 \\
& ours & 20.14 & 0.1009 & \textbf{12.87} & \textbf{0.2261} & \textbf{12.87} & \textbf{0.2261} \\
\midrule
\multirow{3}{*}{shoe, dish, bowl} & baseline & 22.52 & 0.0722 & 23.57 & 0.0656 & 24.44 & 0.0580 \\
& naive-unlearning & 21.78 & 0.0813 & 21.28 & 0.0933 & 23.30 & 0.0693 \\
& ours & 19.52 & 0.1037 & \textbf{12.62} & \textbf{0.2311} & \textbf{12.62} & \textbf{0.2311} \\
\bottomrule
\end{tabular}
}
\end{table}

\begin{table}[h]
\centering
\caption{Ablation study on the combination of multiple attribute traps (GSO $\rightarrow$ Omni). The results highlight that integrating multiple traps significantly enhances the stability of the source domain performance. Abbreviations op, sc, pos, cl, and rt denote opacity, scale, position, color, and rotation traps respectively.  Target metrics are recorded at 200 and 400 steps.}
\label{tab:ablation_combination}
\resizebox{\columnwidth}{!}{
\begin{tabular}{l|l|cc|cc|cc}
\toprule
\multirow{2}{*}{\textbf{Category}} & \multirow{2}{*}{\textbf{Trap Combination}} & \multicolumn{2}{c|}{\textbf{Source Utility Avg.}} & \multicolumn{2}{c|}{\textbf{Target 200 Steps}} & \multicolumn{2}{c}{\textbf{Target 400 Steps}} \\
\cmidrule(lr){3-4} \cmidrule(lr){5-6} \cmidrule(lr){7-8}
& & PSNR $\uparrow$ & LPIPS $\downarrow$ & PSNR $\downarrow$ & LPIPS $\uparrow$ & PSNR $\downarrow$ & LPIPS $\uparrow$ \\
\midrule
\multirow{6}{*}{shoe} & op & 17.93 & 0.1255 & \textbf{13.71} & \textbf{0.1916} & \textbf{13.71} & \textbf{0.1916} \\
& sc & 19.74 & 0.1333 & 20.68 & 0.1233 & 23.43 & 0.0656 \\
& op + sc & 19.62 & 0.1053 & \textbf{13.71} & \textbf{0.1916} & \textbf{13.71} & \textbf{0.1916} \\
& op + sc + pos & 20.06 & 0.0999 & \textbf{13.71} & \textbf{0.1916} & \textbf{13.71} & \textbf{0.1916} \\
& op + sc + pos + rt & 20.28 & 0.0982 & 14.10 & 0.1825 & 21.81 & 0.0777 \\
& op + sc + pos + rt + cl & \textbf{20.67} & \textbf{0.0890} & \textbf{13.71} & 0.1915 & 13.76 & 0.1900 \\
\midrule
\multirow{6}{*}{plant} & op & 18.95 & 0.1117 & \textbf{10.82} & \textbf{0.2256} & \textbf{15.36} & \textbf{0.1601} \\
& sc & 18.44 & 0.1670 & 18.94 & 0.1582 & 21.96 & 0.0916 \\
& op + sc & 20.76 & 0.0915 & 21.07 & 0.1092 & 22.53 & 0.0788 \\
& op + sc + pos & 21.77 & 0.0801 & 19.84 & 0.1177 & 21.86 & 0.0829 \\
& op + sc + pos + rt & 22.22 & 0.0751 & 20.89 & 0.0996 & 22.39 & 0.0779 \\
& op + sc + pos + rt + cl & \textbf{22.52} & \textbf{0.0745} & 20.69 & 0.1032 & 22.26 & 0.0825 \\
\bottomrule
\end{tabular}
}
\end{table}

\subsubsection{Scalability to Multi-Category Protection}
\label{subsubsec:multi_category}

To demonstrate scalability, we evaluate GaussLock on protecting multiple object categories simultaneously within a single model. As shown in Table \ref{tab:ablation_multi_cat}, GaussLock maintains a consistent defense when protecting a mixture of diverse geometric structures. For a joint setting of bowl and shoe, our method restricts the target PSNR to a persistently low range while the naive-unlearning approach allows performance recovery. In a more complex three-category setting including shoe, dish, and bowl, GaussLock demonstrates reliable stability by keeping the target PSNR at a steady low plateau. Crucially, the average source PSNR remains consistently high. This confirms that our attribute-aware traps scale effectively to multi-concept protection without losing defensive strength or causing catastrophic forgetting on the authorized source domains.

\begin{table}[h]
\centering
\caption{Ablation study on individual attribute traps (GSO $\rightarrow$ Omni). We isolate the defensive effect of each attribute-aware trap and report the target domain performance at 200 and 400 steps.}
\label{tab:ablation_single_trap}
\resizebox{\columnwidth}{!}{
\begin{tabular}{l|l|cc|cc|cc}
\toprule
\multirow{2}{*}{\textbf{Category}} & \multirow{2}{*}{\textbf{Trap Attribute}} & \multicolumn{2}{c|}{\textbf{Source Utility Avg.}} & \multicolumn{2}{c|}{\textbf{Target 200 Steps}} & \multicolumn{2}{c}{\textbf{Target 400 Steps}} \\
\cmidrule(lr){3-4} \cmidrule(lr){5-6} \cmidrule(lr){7-8}
& & PSNR $\uparrow$ & LPIPS $\downarrow$ & PSNR $\downarrow$ & LPIPS $\uparrow$ & PSNR $\downarrow$ & LPIPS $\uparrow$ \\
\midrule
\multirow{6}{*}{shoe} & baseline attack & 22.52 & 0.0722 & 23.96 & 0.0515 & 25.07 & 0.0451 \\
& rotation & \textbf{22.28} & \textbf{0.0738} & 23.84 & 0.0576 & 24.90 & 0.0490 \\
& position & 19.87 & 0.0929 & 19.12 & 0.0982 & 22.09 & 0.0677 \\
& opacity & 17.93 & 0.1255 & \textbf{13.71} & \textbf{0.1916} & \textbf{13.71} & \textbf{0.1916} \\
& color & 21.85 & 0.0784 & 23.20 & 0.0648 & 24.47 & 0.0555 \\
& scale & 19.74 & 0.1333 & 20.68 & 0.1233 & 23.43 & 0.0656 \\
\midrule
\multirow{6}{*}{plant} & baseline attack & 22.52 & 0.0722 & 23.27 & 0.0674 & 24.30 & 0.0593 \\
& rotation & \textbf{22.15} & \textbf{0.0755} & 22.35 & 0.0798 & 23.41 & 0.0690 \\
& position & 19.70 & 0.0965 & 16.57 & 0.1350 & 19.02 & 0.1066 \\
& opacity & 18.95 & 0.1117 & \textbf{10.82} & \textbf{0.2256} & \textbf{15.36} & \textbf{0.1601} \\
& color & 21.90 & 0.0760 & 22.03 & 0.0848 & 23.64 & 0.0756 \\
& scale & 18.44 & 0.1670 & 18.94 & 0.1582 & 21.96 & 0.0916 \\
\bottomrule
\end{tabular}
}
\end{table}


\subsubsection{Ablation on Individual Attribute Traps}
\label{subsubsec:single_trap}

To investigate the defensive contribution of each physical attribute, we conduct an ablation study activating only one specific trap at a time. As shown in Table \ref{tab:ablation_single_trap}, variants targeting position, scale, rotation, opacity, and color confirm that all five designed traps effectively disrupt unauthorized fine-tuning. We consistently observe that opacity and scale traps are particularly potent. The opacity trap triggers a rapid visibility collapse, while the scale trap induces severe structural distortion by forcing an anisotropic collapse of the Gaussian primitives. The position trap also contributes significantly to geometric degradation. Although rotation and color traps successfully disrupt target fidelity, they are relatively less catastrophic in isolation. These results collectively validate that manipulating any individual physical attribute provides a viable and independent defensive vector against malicious adaptation.

\subsubsection{Ablation on Multiple Attribute Traps}
\label{subsubsec:multiple_traps}

Building upon the individual attribute analysis, we further evaluate the effectiveness of combining multiple physical traps. As demonstrated in Table \ref{tab:ablation_combination}, deploying potent isolated constraints like the opacity or scale trap easily leads to a collapse in authorized source utility. By integrating multiple traps, our framework effectively distributes the defensive burden across different physical attributes. For instance, pairing opacity and scale traps mitigates the aggressive source degradation caused by either trap alone. This composite strategy significantly improves generative fidelity on the source domain while maintaining considerable suppression against unauthorized target fine-tuning. Consequently, this customizable synergy provides a highly adaptable and stable security solution.

\subsubsection{Qualitative Visualization}
\label{subsec:Qualitative Visualization}

\begin{figure}[!t]
\centering
\includegraphics[width=\linewidth]{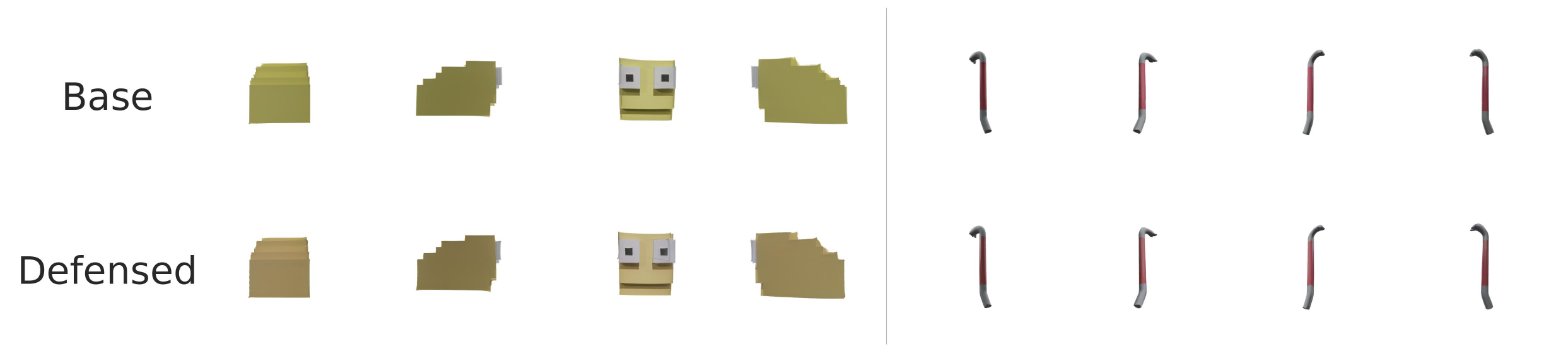}
\caption{Visualization of authorized domain utility preservation. \textbf{Top row:} High-quality generation from the original unprotected weights. \textbf{Bottom row:} Clear and high-fidelity generation from our immunized weights, showing almost no perceptible degradation on authorized task capabilities.}
\label{fig:SourcePreservation}
\end{figure}

\begin{figure}[!t]
\centering
\includegraphics[width=\linewidth]{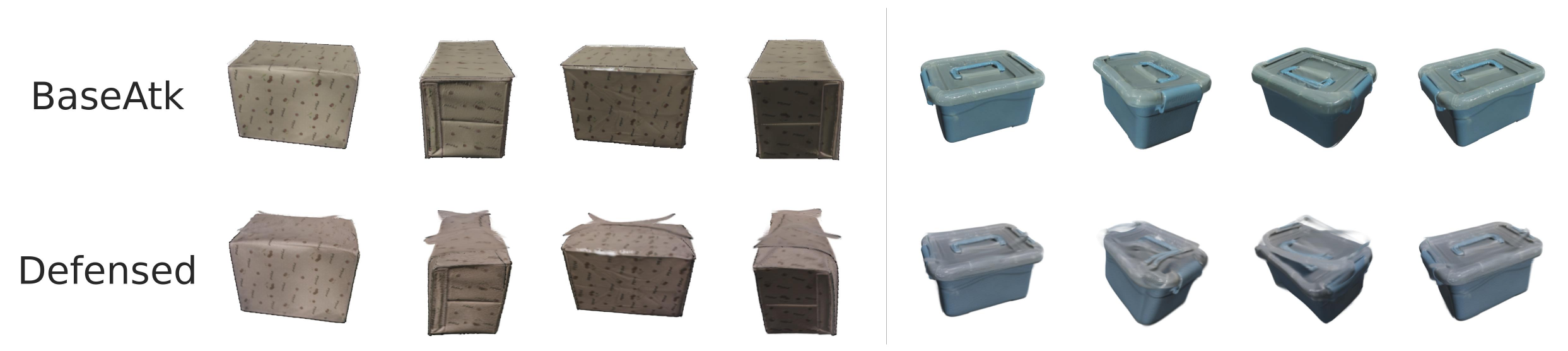}
\caption{Visualization of LoRA fine-tuning attack results on the unauthorized domain. \textbf{Top row:} Unprotected baseline showing successful unauthorized generation. \textbf{Bottom row:} Our GaussLock model showing successful defense via structural collapse.}
\label{fig:Visualization}
\end{figure}

\begin{figure}[!t]
\centering
\includegraphics[width=\linewidth]{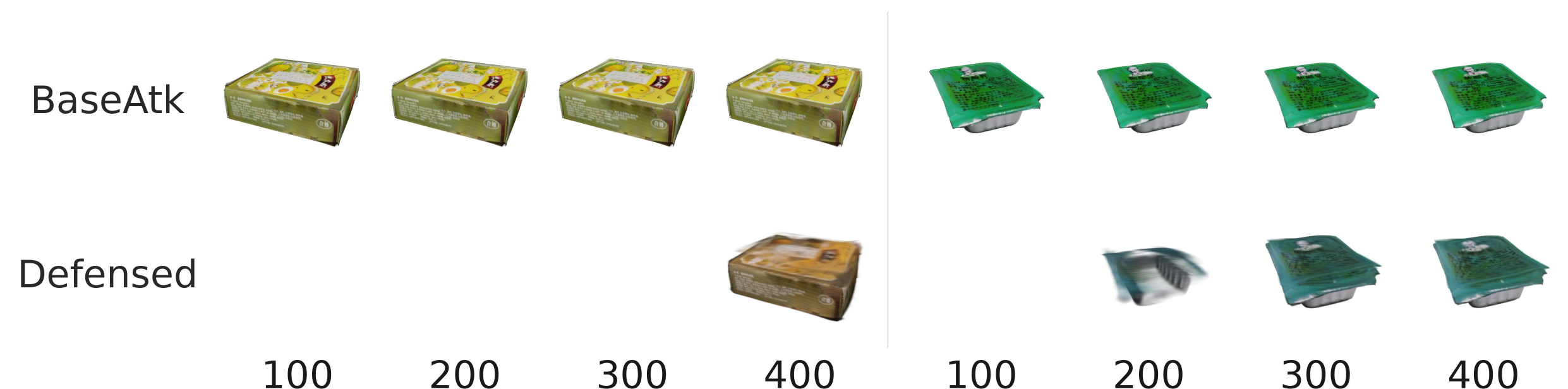}
\caption{Visual comparison of target domain generation under continuous LoRA fine-tuning attacks from 100 to 400 steps (GSO $\rightarrow$ Omni). \textbf{Top row:} Unprotected baseline attack progression. \textbf{Bottom row:} Generation under our GaussLock defense. Low step counts result in complete transparency, while high step counts produce only blurry and degraded objects, confirming persistent suppression throughout the attack.}
\label{fig:TargetEvolution}
\end{figure}

As shown in Fig. \ref{fig:SourcePreservation} and Fig. \ref{fig:Visualization}, our immunized weights rigorously preserve authorized fidelity while inducing severe structural collapse during unauthorized LoRA attacks. To qualitatively evaluate GaussLock, we analyze the visual outcomes of the target domain generation under continuous fine-tuning attacks. Fig. \ref{fig:TargetEvolution} illustrates these visual comparisons across optimization steps ranging from 100 to 400. While the unprotected baseline steadily converges to yield high-quality 3D objects, our protected model demonstrates persistent structural suppression. The generated views initially exhibit complete transparency due to the opacity trap. As the optimization continues into higher step counts, the geometry fails to recover and instead evolves into blurry and degraded shapes. This visual evidence confirms that our protection mechanism directly disrupts the explicit 3D Gaussian parameters. It effectively diminishes the visual utility of the stolen asset and maintains a robust defensive state throughout the entire adaptation process.

\section{Conclusion}
\label{sec:conclusion}

In this paper, we presented GaussLock, a lightweight and proactive framework defending 3D generative models against unauthorized fine-tuning attacks. By embedding dormant attribute-aware traps directly into the explicit 3D Gaussian parameter space, our approach streamlines optimization by bypassing the differentiable rendering pipeline while ensuring efficient source distillation. Extensive evaluations demonstrate that GaussLock induces an irreversible structural collapse in unauthorized target domains across various attack configurations while strictly preserving generation fidelity on the original source tasks. The demonstrated robustness in both intra-domain and cross-domain scenarios confirms that these parameter-level physical traps provide a formidable defense against intellectual property theft. Ultimately, GaussLock offers a scalable paradigm for securing explicit 3D representations within the rapidly evolving generative artificial intelligence ecosystem. Future work will explore extending this parameter-space immunization to other 3D representations and investigating more adaptive trap
  mechanisms to counter emerging adversarial techniques.





\bibliographystyle{IEEEtran}
\bibliography{main}

\begin{thebibliography}{10}
\providecommand{\url}[1]{#1}
\csname url@samestyle\endcsname
\providecommand{\newblock}{\relax}
\providecommand{\bibinfo}[2]{#2}
\providecommand{\BIBentrySTDinterwordspacing}{\spaceskip=0pt\relax}
\providecommand{\BIBentryALTinterwordstretchfactor}{4}
\providecommand{\BIBentryALTinterwordspacing}{\spaceskip=\fontdimen2\font plus
\BIBentryALTinterwordstretchfactor\fontdimen3\font minus \fontdimen4\font\relax}
\providecommand{\BIBforeignlanguage}[2]{{%
\expandafter\ifx\csname l@#1\endcsname\relax
\typeout{** WARNING: IEEEtran.bst: No hyphenation pattern has been}%
\typeout{** loaded for the language `#1'. Using the pattern for}%
\typeout{** the default language instead.}%
\else
\language=\csname l@#1\endcsname
\fi
#2}}
\providecommand{\BIBdecl}{\relax}
\BIBdecl

\bibitem{achlioptas2018learning}
P.~Achlioptas, O.~Diamanti, I.~Mitliagkas, and L.~Guibas, ``Learning representations and generative models for 3d point clouds,'' in \emph{International conference on machine learning}.\hskip 1em plus 0.5em minus 0.4em\relax PMLR, 2018, pp. 40--49.

\bibitem{chaudhuri2020learning}
S.~Chaudhuri, D.~Ritchie, J.~Wu, K.~Xu, and H.~Zhang, ``Learning generative models of 3d structures,'' in \emph{Computer graphics forum}, vol.~39, no.~2.\hskip 1em plus 0.5em minus 0.4em\relax Wiley Online Library, 2020, pp. 643--666.

\bibitem{gao2022get3d}
J.~Gao, T.~Shen, Z.~Wang, W.~Chen, K.~Yin, D.~Li, O.~Litany, Z.~Gojcic, and S.~Fidler, ``Get3d: A generative model of high quality 3d textured shapes learned from images,'' \emph{Advances in neural information processing systems}, vol.~35, pp. 31\,841--31\,854, 2022.

\bibitem{geiger2011stereoscan}
A.~Geiger, J.~Ziegler, and C.~Stiller, ``Stereoscan: Dense 3d reconstruction in real-time,'' in \emph{2011 IEEE intelligent vehicles symposium (IV)}.\hskip 1em plus 0.5em minus 0.4em\relax Ieee, 2011, pp. 963--968.

\bibitem{sayed2022simplerecon}
M.~Sayed, J.~Gibson, J.~Watson, V.~Prisacariu, M.~Firman, and C.~Godard, ``Simplerecon: 3d reconstruction without 3d convolutions,'' in \emph{European Conference on Computer Vision}.\hskip 1em plus 0.5em minus 0.4em\relax Springer, 2022, pp. 1--19.

\bibitem{shi2022deep}
Z.~Shi, S.~Peng, Y.~Xu, A.~Geiger, Y.~Liao, and Y.~Shen, ``Deep generative models on 3d representations: A survey,'' \emph{arXiv preprint arXiv:2210.15663}, 2022.

\bibitem{wang2023rodin}
T.~Wang, B.~Zhang, T.~Zhang, S.~Gu, J.~Bao, T.~Baltrusaitis, J.~Shen, D.~Chen, F.~Wen, Q.~Chen \emph{et~al.}, ``Rodin: A generative model for sculpting 3d digital avatars using diffusion,'' in \emph{Proceedings of the IEEE/CVF conference on computer vision and pattern recognition}, 2023, pp. 4563--4573.

\bibitem{xu2024grm}
Y.~Xu, Z.~Shi, W.~Yifan, H.~Chen, C.~Yang, S.~Peng, Y.~Shen, and G.~Wetzstein, ``Grm: Large gaussian reconstruction model for efficient 3d reconstruction and generation,'' in \emph{European Conference on Computer Vision}.\hskip 1em plus 0.5em minus 0.4em\relax Springer, 2024, pp. 1--20.

\bibitem{wen20233d}
C.~Wen, X.~Li, H.~Huang, Y.-S. Liu, and Y.~Fang, ``3d shape contrastive representation learning with adversarial examples,'' \emph{IEEE Transactions on Multimedia}, vol.~27, pp. 679--692, 2023.

\bibitem{zheng2026llava}
P.~Zheng, C.~Zhang, J.~Mo, G.~Li, J.~Zhang, J.~Zhang, S.~Cao, S.~Zheng, C.~Qin, G.~Wang \emph{et~al.}, ``Llava-fa: Learning fourier approximation for compressing large multimodal models,'' \emph{arXiv preprint arXiv:2602.00135}, 2026.

\bibitem{buyuksalih20173d}
I.~Buyuksalih, S.~Bayburt, G.~Buyuksalih, A.~Baskaraca, H.~Karim, and A.~A. Rahman, ``3d modelling and visualization based on the unity game engine--advantages and challenges,'' \emph{ISPRS Annals of the Photogrammetry, Remote Sensing and Spatial Information Sciences}, vol.~4, pp. 161--166, 2017.

\bibitem{cellary2012interactive}
W.~Cellary and K.~Walczak, \emph{Interactive 3D multimedia content}.\hskip 1em plus 0.5em minus 0.4em\relax Springer, 2012.

\bibitem{zhang2025purifier}
X.~Zhang, Y.~Jin, H.~Tong, J.~Lou, K.~Wu, and X.~Chen, ``Purifier $^{+}$: Plug-and-play backdoor mitigation for pre-trained models via activation alignment,'' \emph{IEEE Transactions on Multimedia}, vol.~27, pp. 3910--3924, 2025.

\bibitem{chattopadhyay2025one}
T.~Chattopadhyay, F.~Ceschin, M.~E. Garza, D.~Zyunkin, A.~Chhotaray, A.~P. Stebner, S.~Zonouz, and R.~Beyah, ``One video to steal them all: 3d-printing ip theft through optical side-channels,'' in \emph{Proceedings of the 2025 ACM SIGSAC Conference on Computer and Communications Security}, 2025, pp. 723--737.

\bibitem{dolgavin2025turning}
A.~Dolgavin, J.~Gatlin, M.~Yung, and M.~Yampolskiy, ``Turning hearsay into discovery: Industrial 3d printer side channel information translated to stealing the object design,'' \emph{arXiv preprint arXiv:2509.18366}, 2025.

\bibitem{zheng2025joint}
P.~Zheng, X.~Pu, K.~Chen, J.~Huang, M.~Yang, B.~Feng, Y.~Ren, J.~Jiang, C.~Zhang, Y.~Yang \emph{et~al.}, ``Joint lossless compression and steganography for medical images via large language models,'' \emph{arXiv preprint arXiv:2508.01782}, 2025.

\bibitem{mildenhall2021nerf}
B.~Mildenhall, P.~P. Srinivasan, M.~Tancik, J.~T. Barron, R.~Ramamoorthi, and R.~Ng, ``Nerf: Representing scenes as neural radiance fields for view synthesis,'' \emph{Communications of the ACM}, vol.~65, no.~1, pp. 99--106, 2021.

\bibitem{schwarz2020graf}
K.~Schwarz, Y.~Liao, M.~Niemeyer, and A.~Geiger, ``Graf: Generative radiance fields for 3d-aware image synthesis,'' \emph{Advances in neural information processing systems}, vol.~33, pp. 20\,154--20\,166, 2020.

\bibitem{zhang2026learning}
J.~Zhang, C.~Zhang, S.~Chen, X.~Wang, Z.~Huang, P.~Zheng, S.~Yuan, S.~Zheng, Q.~Sun, J.~Zou \emph{et~al.}, ``Learning global hypothesis space for enhancing synergistic reasoning chain,'' \emph{arXiv preprint arXiv:2602.09794}, 2026.

\bibitem{qin2024langsplat}
M.~Qin, W.~Li, J.~Zhou, H.~Wang, and H.~Pfister, ``Langsplat: 3d language gaussian splatting,'' in \emph{Proceedings of the IEEE/CVF Conference on Computer Vision and Pattern Recognition}, 2024, pp. 20\,051--20\,060.

\bibitem{yi2024gaussiandreamer}
T.~Yi, J.~Fang, J.~Wang, G.~Wu, L.~Xie, X.~Zhang, W.~Liu, Q.~Tian, and X.~Wang, ``Gaussiandreamer: Fast generation from text to 3d gaussians by bridging 2d and 3d diffusion models,'' in \emph{Proceedings of the IEEE/CVF conference on computer vision and pattern recognition}, 2024, pp. 6796--6807.

\bibitem{ren2024l4gm}
J.~Ren, K.~Xie, A.~Mirzaei, H.~Liang, X.~Zeng, K.~Kreis, Z.~Liu, A.~Torralba, S.~Fidler, S.~W. Kim \emph{et~al.}, ``L4gm: Large 4d gaussian reconstruction model,'' \emph{Advances in Neural Information Processing Systems}, vol.~37, pp. 56\,828--56\,858, 2024.

\bibitem{zhu2025large}
L.~Zhu, G.~Lin, J.~Chen, X.~Zhang, Z.~Jin, Z.~Wang, and L.~Yu, ``Large images are gaussians: High-quality large image representation with levels of 2d gaussian splatting,'' in \emph{Proceedings of the AAAI Conference on Artificial Intelligence}, vol.~39, no.~10, 2025, pp. 10\,977--10\,985.

\bibitem{tang2024lgm}
J.~Tang, Z.~Chen, X.~Chen, T.~Wang, G.~Zeng, and Z.~Liu, ``Lgm: Large multi-view gaussian model for high-resolution 3d content creation,'' in \emph{European Conference on Computer Vision}.\hskip 1em plus 0.5em minus 0.4em\relax Springer, 2024, pp. 1--18.

\bibitem{wang2025f3d}
Y.~Wang, Q.~Wu, and D.~Xu, ``F3d-gaus: Feed-forward 3d-aware generation on imagenet with cycle-aggregative gaussian splatting,'' \emph{arXiv preprint arXiv:2501.06714}, 2025.

\bibitem{yang2025prometheus}
Y.~Yang, J.~Shao, X.~Li, Y.~Shen, A.~Geiger, and Y.~Liao, ``Prometheus: 3d-aware latent diffusion models for feed-forward text-to-3d scene generation,'' in \emph{Proceedings of the IEEE/CVF Conference on Computer Vision and Pattern Recognition}, 2025, pp. 2857--2869.

\bibitem{chen2024pref3r}
Z.~Chen, J.~Yang, and H.~Yang, ``Pref3r: Pose-free feed-forward 3d gaussian splatting from variable-length image sequence,'' \emph{arXiv preprint arXiv:2411.16877}, 2024.

\bibitem{zheng2025lightweight}
P.~Zheng, K.~Chen, J.~Huang, B.~Chen, J.~Liu, Y.~Ren, and X.~Pu, ``Lightweight medical image restoration via integrating reliable lesion-semantic driven prior,'' in \emph{Proceedings of the 33rd ACM International Conference on Multimedia}, 2025, pp. 2977--2986.

\bibitem{sha2022fine}
Z.~Sha, X.~He, P.~Berrang, M.~Humbert, and Y.~Zhang, ``Fine-tuning is all you need to mitigate backdoor attacks,'' \emph{arXiv preprint arXiv:2212.09067}, 2022.

\bibitem{wang2021non}
L.~Wang, S.~Xu, R.~Xu, X.~Wang, and Q.~Zhu, ``Non-transferable learning: A new approach for model ownership verification and applicability authorization,'' \emph{arXiv preprint arXiv:2106.06916}, 2021.

\bibitem{deng2024sophonnonfinetunablelearningrestrain}
\BIBentryALTinterwordspacing
J.~Deng, S.~Pang, Y.~Chen, L.~Xia, Y.~Bai, H.~Weng, and W.~Xu, ``Sophon: Non-fine-tunable learning to restrain task transferability for pre-trained models,'' 2024. [Online]. Available: \url{https://arxiv.org/abs/2404.12699}
\BIBentrySTDinterwordspacing

\bibitem{hong2025toward}
Z.~Hong, Y.~Xiang, and T.~Liu, ``Toward robust non-transferable learning: a survey and benchmark,'' in \emph{Proceedings of the Thirty-Fourth International Joint Conference on Artificial Intelligence}, 2025, pp. 10\,455--10\,463.

\bibitem{rosati2024representationnoisingdefencemechanism}
\BIBentryALTinterwordspacing
D.~Rosati, J.~Wehner, K.~Williams, Łukasz Bartoszcze, D.~Atanasov, R.~Gonzales, S.~Majumdar, C.~Maple, H.~Sajjad, and F.~Rudzicz, ``Representation noising: A defence mechanism against harmful finetuning,'' 2024. [Online]. Available: \url{https://arxiv.org/abs/2405.14577}
\BIBentrySTDinterwordspacing

\bibitem{chen2025sddselfdegradeddefensemalicious}
\BIBentryALTinterwordspacing
Z.~Chen, W.~Lu, X.~Lin, and Z.~Zeng, ``Sdd: Self-degraded defense against malicious fine-tuning,'' 2025. [Online]. Available: \url{https://arxiv.org/abs/2507.21182}
\BIBentrySTDinterwordspacing

\bibitem{huang2024harmful}
T.~Huang, S.~Hu, F.~Ilhan, S.~F. Tekin, and L.~Liu, ``Harmful fine-tuning attacks and defenses for large language models: A survey,'' \emph{arXiv preprint arXiv:2409.18169}, 2024.

\bibitem{nichol2022pointegenerating3dpoint}
\BIBentryALTinterwordspacing
A.~Nichol, H.~Jun, P.~Dhariwal, P.~Mishkin, and M.~Chen, ``Point-e: A system for generating 3d point clouds from complex prompts,'' 2022. [Online]. Available: \url{https://arxiv.org/abs/2212.08751}
\BIBentrySTDinterwordspacing

\bibitem{chen2025harnessing}
Y.~Chen, S.~Zhao, L.~Duan, C.~Ding, and D.~Tao, ``Harnessing text-to-image diffusion models for point cloud self-supervised learning,'' in \emph{Proceedings of the IEEE/CVF International Conference on Computer Vision}, 2025, pp. 26\,156--26\,166.

\bibitem{zhou2025recurrent}
Y.~Zhou, D.~Ye, H.~Zhang, X.~Xu, H.~Sun, Y.~Xu, X.~Liu, and Y.~Zhou, ``Recurrent diffusion for 3d point cloud generation from a single image,'' \emph{IEEE Transactions on Image Processing}, 2025.

\bibitem{bastico2025rethinkingmetricsdiffusionarchitecture}
\BIBentryALTinterwordspacing
M.~Bastico, D.~Ryckelynck, L.~Corté, Y.~Tillier, and E.~Decencière, ``Rethinking metrics and diffusion architecture for 3d point cloud generation,'' 2025. [Online]. Available: \url{https://arxiv.org/abs/2511.05308}
\BIBentrySTDinterwordspacing

\bibitem{liu2023robust}
D.~Liu, W.~Hu, and X.~Li, ``Robust geometry-dependent attack for 3d point clouds,'' \emph{IEEE Transactions on Multimedia}, vol.~26, pp. 2866--2877, 2023.

\bibitem{yao2025adversarial}
R.~Yao, A.~Zhang, Y.~Zhou, J.~Zhao, B.~Liu, and A.~El~Saddik, ``Adversarial geometric attacks for 3d point cloud object tracking,'' \emph{IEEE Transactions on Multimedia}, 2025.

\bibitem{zheng2023cgc}
P.~Zheng, J.~Jiang, Y.~Zhang, C.~Zeng, C.~Qin, and Z.~Li, ``Cgc-net: A context-guided constrained network for remote-sensing image super resolution,'' \emph{Remote Sensing}, vol.~15, no.~12, p. 3171, 2023.

\bibitem{zhang2024jointly}
Y.~Zhang, P.~Zheng, C.~Zeng, B.~Xiao, Z.~Li, and X.~Gao, ``Jointly rs image deblurring and super-resolution with adjustable-kernel and multi-domain attention,'' \emph{IEEE Transactions on Geoscience and Remote Sensing}, vol.~63, pp. 1--16, 2024.

\bibitem{zheng2026towards}
P.~Zheng, C.~Zhang, M.~Cui, G.~Chen, Q.~Sun, J.~Huang, J.~Zhang, T.-H. Kim, C.~Qin, Y.~Ren \emph{et~al.}, ``Towards visual chain-of-thought reasoning: A comprehensive survey,'' 2026.

\bibitem{kerbl20233d}
B.~Kerbl, G.~Kopanas, T.~Leimk{\"u}hler, G.~Drettakis \emph{et~al.}, ``3d gaussian splatting for real-time radiance field rendering.'' \emph{ACM Trans. Graph.}, vol.~42, no.~4, pp. 139--1, 2023.

\bibitem{liu2020dlgan}
C.~Liu, D.~Kong, S.~Wang, J.~Li, and B.~Yin, ``Dlgan: Depth-preserving latent generative adversarial network for 3d reconstruction,'' \emph{IEEE Transactions on Multimedia}, vol.~23, pp. 2843--2856, 2020.

\bibitem{uchida2017embedding}
Y.~Uchida, Y.~Nagai, S.~Sakazawa, and S.~Satoh, ``Embedding watermarks into deep neural networks,'' in \emph{Proceedings of the 2017 ACM on international conference on multimedia retrieval}, 2017, pp. 269--277.

\bibitem{adi2018turning}
Y.~Adi, C.~Baum, M.~Cisse, B.~Pinkas, and J.~Keshet, ``Turning your weakness into a strength: Watermarking deep neural networks by backdooring,'' in \emph{27th USENIX security symposium (USENIX Security 18)}, 2018, pp. 1615--1631.

\bibitem{lukas2019deep}
N.~Lukas, Y.~Zhang, and F.~Kerschbaum, ``Deep neural network fingerprinting by conferrable adversarial examples,'' \emph{arXiv preprint arXiv:1912.00888}, 2019.

\bibitem{wu2025robust}
S.~Wu, W.~Lu, and X.~Luo, ``Robust watermarking based on multi-layer watermark feature fusion,'' \emph{IEEE Transactions on Multimedia}, 2025.

\bibitem{you2024two}
J.~You and Y.~Zhou, ``Two-stage watermark removal framework for spread spectrum watermarking,'' \emph{IEEE Transactions on Multimedia}, vol.~26, pp. 7687--7699, 2024.

\bibitem{zhong2020automated}
X.~Zhong, P.-C. Huang, S.~Mastorakis, and F.~Y. Shih, ``An automated and robust image watermarking scheme based on deep neural networks,'' \emph{IEEE Transactions on Multimedia}, vol.~23, pp. 1951--1961, 2020.

\bibitem{luo2025diffw}
T.~Luo, R.~Hu, Z.~He, G.~Jiang, H.~Xu, Y.~Song, and C.-C. Chang, ``Diffw: Multi-encoder based on conditional diffusion model for robust image watermarking,'' \emph{IEEE Transactions on Multimedia}, vol.~28, pp. 837--852, 2025.

\bibitem{sun2026grasp}
Q.~Sun, C.~Zhang, J.~Zhang, X.~Wang, J.~Xie, P.~Zheng, H.~Wang, S.~Lee, C.-l.~A. Tai, Y.~Yang \emph{et~al.}, ``Grasp: Guided region-aware sparse prompting for adapting mllms to remote sensing,'' \emph{arXiv preprint arXiv:2601.17089}, 2026.

\bibitem{zhang2026ghs}
J.~Zhang, C.~Zhang, S.~Chen, X.~Wang, Z.~Huang, P.~Zheng, S.~Yuan, S.~Zheng, Q.~Sun, J.~Zou \emph{et~al.}, ``Ghs-tda: A synergistic reasoning framework integrating global hypothesis space with topological data analysis,'' \emph{arXiv e-prints}, pp. arXiv--2602, 2026.

\bibitem{cao2026language}
S.~Cao, J.~Zhang, P.~Zheng, J.~Yan, C.~Qin, Y.~Ye, W.~Dong, P.~Wang, Y.~Yang, and C.~Zhang, ``Language-guided token compression with reinforcement learning in large vision-language models,'' \emph{arXiv preprint arXiv:2603.13394}, 2026.

\bibitem{shan2023glaze}
S.~Shan, J.~Cryan, E.~Wenger, H.~Zheng, R.~Hanocka, and B.~Y. Zhao, ``Glaze: Protecting artists from style mimicry by $\{$Text-to-Image$\}$ models,'' in \emph{32nd USENIX Security Symposium (USENIX Security 23)}, 2023, pp. 2187--2204.

\bibitem{shan2024nightshade}
S.~Shan, W.~Ding, J.~Passananti, S.~Wu, H.~Zheng, and B.~Y. Zhao, ``Nightshade: Prompt-specific poisoning attacks on text-to-image generative models,'' in \emph{2024 IEEE symposium on security and privacy (SP)}.\hskip 1em plus 0.5em minus 0.4em\relax IEEE, 2024, pp. 807--825.

\bibitem{shan2020fawkes}
S.~Shan, E.~Wenger, J.~Zhang, H.~Li, H.~Zheng, and B.~Y. Zhao, ``Fawkes: Protecting privacy against unauthorized deep learning models,'' in \emph{29th USENIX security symposium (USENIX Security 20)}, 2020, pp. 1589--1604.

\bibitem{wan2023poisoning}
A.~Wan, E.~Wallace, S.~Shen, and D.~Klein, ``Poisoning language models during instruction tuning,'' in \emph{International Conference on Machine Learning}.\hskip 1em plus 0.5em minus 0.4em\relax PMLR, 2023, pp. 35\,413--35\,425.

\bibitem{zheng2023targeted}
B.~Zheng, C.~Liang, and X.~Wu, ``Targeted attack improves protection against unauthorized diffusion customization,'' \emph{arXiv preprint arXiv:2310.04687}, 2023.

\bibitem{hong2024improving}
Z.~Hong, Z.~Wang, L.~Shen, Y.~Yao, Z.~Huang, S.~Chen, C.~Yang, M.~Gong, and T.~Liu, ``Improving non-transferable representation learning by harnessing content and style,'' in \emph{The twelfth international conference on learning representations}, 2024.

\bibitem{wang2024say}
L.~Wang, M.~Wang, H.~Fu, and D.~Zhang, ``Say no to freeloader: Protecting intellectual property of your deep model,'' \emph{IEEE Transactions on Pattern Analysis and Machine Intelligence}, vol.~46, no.~12, pp. 11\,073--11\,086, 2024.

\bibitem{wang2026towards}
Z.~Wang, N.~Li, P.~Li, G.~Sun, T.~Chen, and A.~Li, ``Towards building non-fine-tunable foundation models,'' \emph{arXiv preprint arXiv:2602.00446}, 2026.

\bibitem{peng2026dynamic}
B.~Peng, S.~Qu, Y.~Wu, T.~Zou, L.~He, A.~Knoll, G.~Chen, and C.~Jiang, ``Dynamic mask-pruning strategy for source-free model intellectual property protection,'' \emph{International Journal of Computer Vision}, vol. 134, no.~2, p.~56, 2026.

\bibitem{hu2025adaptive}
Z.~Hu, L.~Shen, Z.~Wang, Y.~Wei, and D.~Tao, ``Adaptive defense against harmful fine-tuning for large language models via bayesian data scheduler,'' \emph{arXiv preprint arXiv:2510.27172}, 2025.

\bibitem{huangbooster}
T.~Huang, S.~Hu, F.~Ilhan, S.~F. Tekin, and L.~Liu, ``Booster: Tackling harmful fine-tuning for large language models via attenuating harmful perturbation,'' in \emph{The Thirteenth International Conference on Learning Representations}, 2025.

\bibitem{huang2024vaccine}
T.~Huang, S.~Hu, and L.~Liu, ``Vaccine: Perturbation-aware alignment for large language models against harmful fine-tuning attack,'' \emph{Advances in Neural Information Processing Systems}, vol.~37, pp. 74\,058--74\,088, 2024.

\bibitem{qi2023fine}
X.~Qi, Y.~Zeng, T.~Xie, P.-Y. Chen, R.~Jia, P.~Mittal, and P.~Henderson, ``Fine-tuning aligned language models compromises safety, even when users do not intend to!'' \emph{arXiv preprint arXiv:2310.03693}, 2023.

\bibitem{qisafety}
X.~Qi, A.~Panda, K.~Lyu, X.~Ma, S.~Roy, A.~Beirami, P.~Mittal, and P.~Henderson, ``Safety alignment should be made more than just a few tokens deep,'' in \emph{The Thirteenth International Conference on Learning Representations}, 2025.

\bibitem{liu2019soft}
S.~Liu, T.~Li, W.~Chen, and H.~Li, ``Soft rasterizer: A differentiable renderer for image-based 3d reasoning,'' in \emph{Proceedings of the IEEE/CVF international conference on computer vision}, 2019, pp. 7708--7717.

\bibitem{chen2019learning}
W.~Chen, H.~Ling, J.~Gao, E.~Smith, J.~Lehtinen, A.~Jacobson, and S.~Fidler, ``Learning to predict 3d objects with an interpolation-based differentiable renderer,'' \emph{Advances in neural information processing systems}, vol.~32, 2019.

\bibitem{zhao2026intellectual}
L.~Zhao, Z.~Hong, J.~Huang, R.~Chen, M.~Gong, and T.~Liu, ``Intellectual property protection for 3d gaussian splatting assets: A survey,'' \emph{arXiv preprint arXiv:2602.03878}, 2026.

\bibitem{wang2025machine}
S.~Wang, J.~Ye, and X.~Wang, ``Machine unlearning in 3d generation: A perspective-coherent acceleration framework,'' in \emph{The Thirty-ninth Annual Conference on Neural Information Processing Systems}, 2025.

\bibitem{hong2025adlift}
Z.~Hong, T.~Huang, R.~Chen, S.~Ye, M.~Gong, B.~Han, and T.~Liu, ``Adlift: Lifting adversarial perturbations to safeguard 3d gaussian splatting assets against instruction-driven editing,'' \emph{arXiv preprint arXiv:2512.07247}, 2025.

\bibitem{zhao2025rdsplat}
L.~Zhao, Z.~Hong, Z.~Ren, R.~Chen, M.~Gong, and T.~Liu, ``Rdsplat: Robust watermarking against diffusion editing for 3d gaussian splatting,'' \emph{arXiv preprint arXiv:2512.06774}, 2025.

\bibitem{huang2024gaussianmarker}
X.~Huang, R.~Li, Y.-m. Cheung, K.~C. Cheung, S.~See, and R.~Wan, ``Gaussianmarker: Uncertainty-aware copyright protection of 3d gaussian splatting,'' \emph{Advances in Neural Information Processing Systems}, vol.~37, pp. 33\,037--33\,060, 2024.

\bibitem{ronneberger2015u}
O.~Ronneberger, P.~Fischer, and T.~Brox, ``U-net: Convolutional networks for biomedical image segmentation,'' in \emph{International Conference on Medical image computing and computer-assisted intervention}.\hskip 1em plus 0.5em minus 0.4em\relax Springer, 2015, pp. 234--241.

\bibitem{hu2022lora}
E.~J. Hu, Y.~Shen, P.~Wallis, Z.~Allen-Zhu, Y.~Li, S.~Wang, L.~Wang, W.~Chen \emph{et~al.}, ``Lora: Low-rank adaptation of large language models.'' \emph{Iclr}, vol.~1, no.~2, p.~3, 2022.

\bibitem{hore2010image}
A.~Hore and D.~Ziou, ``Image quality metrics: Psnr vs. ssim,'' in \emph{2010 20th international conference on pattern recognition}.\hskip 1em plus 0.5em minus 0.4em\relax IEEE, 2010, pp. 2366--2369.

\bibitem{zhang2018unreasonable}
R.~Zhang, P.~Isola, A.~A. Efros, E.~Shechtman, and O.~Wang, ``The unreasonable effectiveness of deep features as a perceptual metric,'' in \emph{Proceedings of the IEEE conference on computer vision and pattern recognition}, 2018, pp. 586--595.

\bibitem{downs2022google}
L.~Downs, A.~Francis, N.~Koenig, B.~Kinman, R.~Hickman, K.~Reymann, T.~B. McHugh, and V.~Vanhoucke, ``Google scanned objects: A high-quality dataset of 3d scanned household items,'' in \emph{2022 International Conference on Robotics and Automation (ICRA)}.\hskip 1em plus 0.5em minus 0.4em\relax Ieee, 2022, pp. 2553--2560.

\bibitem{wu2023omniobject3d}
T.~Wu, J.~Zhang, X.~Fu, Y.~Wang, J.~Ren, L.~Pan, W.~Wu, L.~Yang, J.~Wang, C.~Qian \emph{et~al.}, ``Omniobject3d: Large-vocabulary 3d object dataset for realistic perception, reconstruction and generation,'' in \emph{Proceedings of the IEEE/CVF conference on computer vision and pattern recognition}, 2023, pp. 803--814.

\end{thebibliography}


\vfill

\end{document}